\documentclass[a4paper, 10pt, conference]{ieeeconf}      % Use this line for a4
\usepackage{FG2024} 

\FGfinalcopy % *** Uncomment this line for the final submission

\IEEEoverridecommandlockouts                              % This command is only
\usepackage{hyperref} 
\usepackage{graphicx}
\usepackage{wrapfig, booktabs}
\usepackage{amsmath, mathtools}
\usepackage{nccmath}
\usepackage{amssymb}
\usepackage{bm}
\usepackage{cuted}
\usepackage{placeins}

\usepackage[subrefformat=parens]{subcaption}

\usepackage{url}
\usepackage{algpseudocode}
\usepackage{algorithm}
\algrenewcommand\algorithmicrequire{\textbf{Input:}}
\algrenewcommand\algorithmicensure{\textbf{Output:}}
\usepackage{diagbox}
\usepackage{xcolor}
\usepackage{multicol,multirow}
\usepackage{tcolorbox}
\usepackage[bottom]{footmisc} % added to resolve footnote problem :)

\usepackage{xspace}
\makeatletter
\DeclareRobustCommand\onedot{\futurelet\@let@token\@onedot}
\def\@onedot{\ifx\@let@token.\else.\null\fi\xspace}
\def\eg{\emph{e.g}\onedot} 
\def\ie{\emph{i.e}\onedot} 
 
\def\etc{\emph{etc}\onedot} 
 
\def\etal{\emph{et al}\onedot}
\makeatother

\def\FGPaperID{267} % *** Enter the FG2024 Paper ID here

\title{\LARGE \bf
Discovering Interpretable Directions in the Semantic Latent Space of Diffusion Models
}

\author{\parbox{16cm}{\centering
    {\large  René Haas$^1$ and Inbar Huberman-Spiegelglas$^2$ and Rotem Mulayoff$^2$ and Stella Graßhof$^1$ and Sami S. Brandt$^1$ and Tomer Michaeli$^2$ }\\
    {\normalsize
    $^1$ Computer Science, IT University of Copenhagen, Denmark\\
    $^2$ Computer Science, Technion, Israel}}
}

\usepackage{fancyhdr}
\begin{document}

\ifFGfinal
\thispagestyle{empty}
\pagestyle{empty}
\else
\author{Anonymous FG2024 submission\\ Paper ID \FGPaperID \\}
\pagestyle{plain}
\fi
\maketitle
\thispagestyle{fancy} 
%%%%%%%%%%%%%%%%%%%%%%%%%%%%%%%%%%%%%%%%%%

\begin{strip}
\centering
\includegraphics[width=\linewidth]{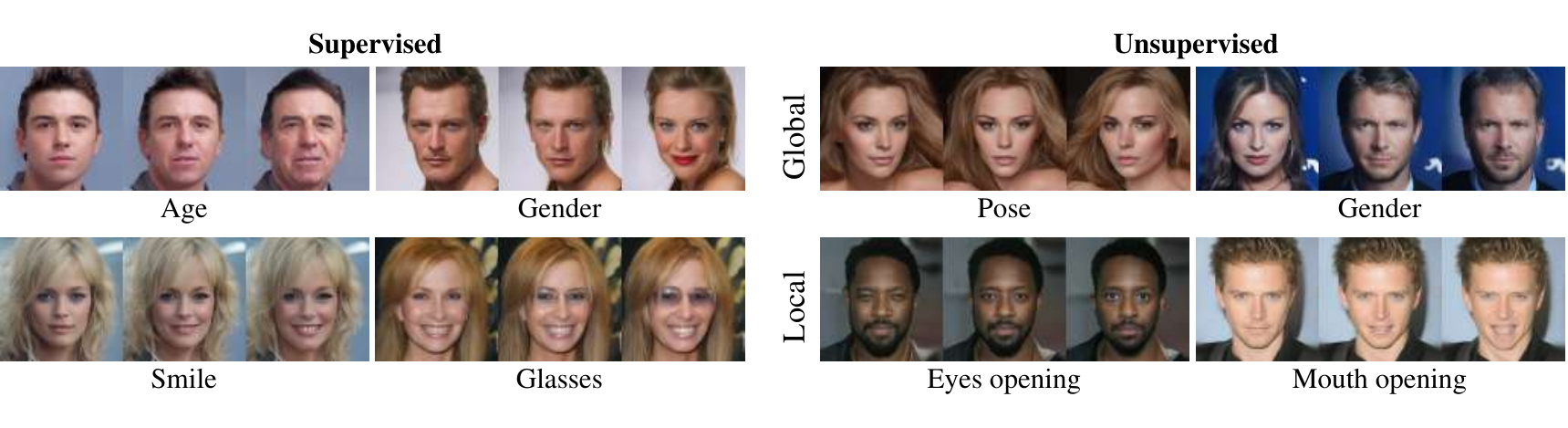}
\captionsetup{type=figure}
\captionof{figure}{
\textbf{Our semantic image editing.} 
We present new methods for finding interpretable disentangled semantic directions in the latent space of DDMs. 
Specifically, we propose a supervised (left) and two unsupervised (right) methods, where the latter finds either global directions based on a collection of images or local directions based on the analysis of a single sample.
}
\label{fig:teaser figure}
\end{strip}

\begin{abstract}
Denoising Diffusion Models (DDMs) have emerged as a strong competitor to Generative Adversarial Networks (GANs).
However, despite their widespread use in image synthesis and editing applications, their latent space is still not as well understood. Recently, a semantic latent space for DDMs, coined `$h$-space', was shown to facilitate semantic image editing in a way reminiscent of GANs. The $h$-space is comprised of the bottleneck activations in the DDM's denoiser across all timesteps of the diffusion process. In this paper, we explore the properties of $h$-space and propose several novel methods for finding meaningful semantic directions within it. We start by studying unsupervised methods for revealing interpretable semantic directions in pretrained DDMs. Specifically, we show that interpretable directions emerge as the principal components in the latent space. Additionally, we provide a novel method for discovering image-specific semantic directions by spectral analysis of the Jacobian of the denoiser w.r.t.\@ the latent code. 
Next, we extend the analysis by finding directions in a supervised fashion in unconditional DDMs.
We demonstrate how such directions can be found by annotating generated samples with a domain-specific attribute classifier. 
We further show how to semantically disentangle the found directions by simple linear projection.
Our approaches are applicable without requiring any architectural modifications, text-based guidance, CLIP-based optimization, or model fine-tuning.
\end{abstract}

%% Moved to first page in main file. 
% \begin{figure*}[bt]
%     \includegraphics[width=\linewidth]{figs/base/introfig/first_figure.pdf}
%   \caption{
%   \textbf{Our semantic image editing.} 
%   We present new methods for finding interpretable disentangled semantic directions in the latent space of DDMs. 
%   Specifically, we propose a supervised (left) and two unsupervised (right) methods, where the latter finds either global directions based on a collection of images or local directions based on the analysis of a single sample.
%   }
%   \label{fig:teaser figure}
% \end{figure*}

\section{Introduction}

% DPMs outperform GANs but their latent space is not well understood
Denoising Diffusion Models (DDMs) \cite{sohl2015ddpm-noneqthermo} have emerged as a strong alternative to Generative Adversarial Networks (GANs) \cite{Goodfellow2014GAN}. Today, they outperform GANs in unconditional image synthesis \cite{Dhariwal2021dpmsbeatgans}, a task in which GANs have been dominating in recent years. Besides synthesizing high-quality and diverse images, DDMs can also be used for conditional synthesis tasks by guiding them on various user inputs \cite{ho2021classifierfree}, such as a user-provided reference image~\cite{Gihyun22, meng2022sdedit} or a text-prompt by utilizing Contrastive Language-Image Pretraining (CLIP) \cite{Radford2021CLIP}. 
Conditional DDMs have seen great success, particularly in the context of text-based synthesis. 
Specifically, recent large-scale text-conditional systems like DALL-E \cite{ramesh2021dalle,ramesh2022dalle2}, Stable Diffusion \cite{rombach2021latentdiffusion} and Imagen \cite{Saharia2022Imagen} have sparked a surge of research related to text-driven image editing using DDMs \cite{nichol2021glide, mokady2022null, gal2022textual, ruiz2022dreambooth, kawar2023imagic, kim2022DiffusionCLIP, Hertz22,Narek22,couairon2022diffedit}. 
While there has been extensive research on finding disentangled editing directions in the latent space of unconditional GANs \cite{alaluf2023third, Shen2020Interfacegan,harkonen2020ganspace,Haas2022tensorGAN2,shen2021closedform, spingarn2021gansteerability,  Abdal2020Image2StyleGANpp}, comparatively little work has been done on this topic for unconditional DDMs. Despite their popularity, it is still not well understood how to leverage the latent space of DDMs for semantic image editing in the unconditional setting, \ie, in the absence of CLIP-guidance and without conditioning on a reference image.

%% Our paper  
In this paper, we propose novel editing techniques by utilizing the \emph{semantic latent space} of DDMs which was recently proposed by Kwon \etal~\cite{Kwon2022ddmhavesemantic}. The semantic latent space, coined `$h$-space', is the space of the deepest feature maps of the denoiser. Our research explores supervised and unsupervised methods for finding semantically interpretable editing directions in unconditional DDMs.

%% Paper Organization
We start by proposing two unsupervised methods. In Sec.~\ref{sec:Unsupervised methods}, we demonstrate that interpretable editing directions, like pose, gender, and age emerge as the principal components in the semantic latent space. Additionally, we propose a novel unsupervised method for discovering image-specific semantic directions resulting in highly localized edits like opening/closing of the mouth and eyes that can also be applied to other samples. 
We illustrate a selection of these unsupervised editing directions in Fig.~\ref{fig:teaser figure} (right pane). 
% Supervised directions. 
Next, in Sec.~\ref{sec:Supervised methods}, we utilize the linear properties of the semantic latent space and propose a simple supervised method for finding interpretable editing directions, like age and gender or the appearance of glasses or a smile. We illustrate examples of these edits in Fig.~\ref{fig:teaser figure} (left pane). 
We demonstrate our approach by annotating samples generated by an unconditional DDM using a pretrained attribute classifier. 
We further propose a simple method for disentangling directions that affect multiple attributes. 
 % Why our approach is attractive, main punchlines.
Our approaches allow for intuitive and semantically disentangled image editing and can be applied to the latent space of DDMs without requiring any CLIP guidance, fine-tuning, optimization or any adaptations to the architecture of existing DDMs. 

To summarize the contributions of this paper are the following: 
\begin{itemize}%[noitemsep]
    %\item finding semantically interpretable editing directions in unconditional DDMs,
    \item We propose an unsupervised method to uncover semantically meaningful directions in the $h$-space by PCA.
    \item Our method successfully identifies image-specific semantically meaningful directions corresponding to highly localized changes.
    \item We demonstrate a supervised approach to obtain latent directions corresponding to well-defined labels.
    \item We propose a conditional manipulation in $h$-space to disentangle semantic directions.
    \item The code for this project is available at \url{https://github.com/renhaa/semantic-diffusion}. 
\end{itemize}

\section{Related work}

\subsection{The latent space of diffusion models}
%% Semantic space is well understood in GANs but not in DDMs
GANs have a well-defined latent space suitable for semantic editing. To which extent DDMs possess such a convenient latent space is still a topic of ongoing research. Here we start by reviewing two approaches for defining a latent space in DDMs that facilitate semantic editing.

%% DDIM view of the latent space
Using DDIM sampling proposed by Song \etal.~\cite{song2020ddim}, the generative process is a deterministic mapping from a Gaussian noise vector~$\mathbf{x}_T\sim \mathcal{N}(\mathbf{0},\mathbf{I})$ to a sampled image~$\mathbf{x}_0$. 
In the DDIM framework, the fully noised image $\mathbf{x}_T$, can be regarded as the latent representation. 
DDIM has the property that fixing~$\mathbf{x}_T$ leads to images with similar high-level features irrespective of the length of the generative process. 
Furthermore, interpolating between two latent codes $\mathbf{x}_{T}^{(1)}$ and~$\mathbf{x}_{T}^{(2)}$ leads to images that vary smoothly between the two corresponding endpoint images, $\mathbf{x}_{0}^{(1)}$ and~$\mathbf{x}_{0}^{(2)}$.

%DDMs Already Have A Semantic Latent Space perspective,
Kwon \etal~\cite{Kwon2022ddmhavesemantic} propose $h$-space for DDMs, the set of bottleneck feature maps of the U-Net \cite{ronneberger2015unet} across all timesteps, $\{\mathbf{h}_T,\ldots,\mathbf{h}_1\}$ as the latent space.  
Each bottleneck feature map $\mathbf{h}_t$ has a lower spatial dimension but more channels than the output image. They show that semantics can be edited by adding offsets $\Delta \mathbf{h}_t$ to the feature maps during the generative process. 
To find editing directions, they use an optimization procedure involving CLIP, where the semantics to be edited are described by text prompts. 
The $h$-space has the following properties: 
(i) a direction $\Delta \mathbf{h}_t$ has the same semantic effect on different samples; 
(ii) the magnitude of $\Delta \mathbf{h}_t$ controls the strength of the edit; 
(iii) $h$-space is additive in the sense that applying a linear combination of different directions where each $\Delta \mathbf{h}_t$ corresponds to a distinct attribute, results in a generated image where all attributes have been changed.

\subsection{Semantic image editing in generative models}
Semantic editing has been widely explored in GANs \cite{Shen2020Interfacegan,harkonen2020ganspace, Haas2022tensorGAN2,shen2021closedform, spingarn2021gansteerability, Patashnik2021styleclip,Abdal2020Image2StyleGANpp,Tewari2020StyleRig, Wu2020StyleSpace}. 
Shen \etal~\cite{Shen2020Interfacegan} used a binary classifier 
to annotate generated samples and trained a SVM to separate classes like pose, age, and gender. The corresponding linear directions in latent space were then defined as the normal vectors of the separating hyper-planes. H{\"a}rk{\"o}nen \etal~\cite{harkonen2020ganspace} found interpretable control directions in pretrained GANs by applying principal components of latent codes to appropriate layers of the generator.
Another line of work \cite{Haas2022tensorGAN2, shen2021closedform, spingarn2021gansteerability, zhu2021lowrankgan} uses various factorization techniques to define meaningful directions in the latent space of GANs.

Semantic image editing has also been shown in DDMs but many existing methods make adaptations to the architecture, employ text-based optimization or model fine-tuning. 
In DiffusionAE~\cite{Preechakul2022DiffusionAutoencoder}, a DDM was trained in conjunction with an image encoder. This enabled attribute manipulation on real images, including modifications of gender, age, and smile, but requires modifying the DDM architecture.
Another line of work includes DiffusionCLIP~\cite{kim2022DiffusionCLIP}, Imagic~\cite{kawar2023imagic}, 
and UniTune~\cite{Valevski22},  combined CLIP-based text guidance with model fine-tuning.
Unlike these methods, our approaches do not require CLIP-based text-guidance nor model fine-tuning and can be applied to existing DDMs without retraining or adapting the architecture. 

We acknowledge as concurrent work the unsupervised method proposed by Park \etal \cite{park2023unsupervised}. 
They perform spectral analysis on the Jacobian of a mapping from pixel space to a reduced $h$-space consisting of the sum-pooled feature map of the bottleneck representation.  
In comparison, our proposed method is able to operate on the full bottleneck representation using power iteration to circumvent the intractable computational cost of calculating the Jacobian explicitly. We further propose to allow for additional region-specific control by calculating the Jacobian with respect to a region of interest, allowing for fine-grained and highly localized semantic editing.

\section{The semantic latent space of DDMs}
Diffusion models are defined in terms of a forward diffusion process that adds increasing amounts of white Gaussian noise to a clean image $\mathbf{x}_0$ in $T$ steps, and a learned reverse process that gradually removes the noise.
During the forward process each noisy image $\mathbf{x}_t$ is generated as
\begin{equation}    
\mathbf{x}_t = \sqrt{\alpha_t}\mathbf{x}_0 + \sqrt{1-\alpha_t}\mathbf{n},
\end{equation}
where $\mathbf{n} \sim \mathcal{N}(\mathbf{0},\mathbf{I})$ and  the noise schedule is defined by~$\{\alpha_t \}$ .
In \cite{song2020ddim}, generating an image from the model is done by first sampling Gaussian noise $\mathbf{x}_T\sim \mathcal{N}(\mathbf{0},\mathbf{I})$, which is then denoised following the approximate reverse diffusion process 
\begin{equation}
\mathbf{x}_{t-1} = 
\sqrt{\alpha_{t-1}} \mathbf{P}_t(\bm{\epsilon}^\theta_t(\mathbf{x}_t)) 
+ \mathbf{D}_t (\bm{\epsilon}^\theta_t(  \mathbf{x}_t))
+ \sigma_t \mathbf{z}_t, 
\label{eq:ddim-reverse}
\end{equation}
where $\mathbf{z}_t\sim\mathcal{N}(\mathbf{0},\mathbf{I})$. Here $\bm{\epsilon}^\theta_t$ is a neural network (usually a U-Net~\cite{ronneberger2015unet}), which is trained to predict $\mathbf{n}$ from $\mathbf{x}_t$, and the terms 
\begin{align}
\mathbf{P}_t(\bm{\epsilon}^\theta_t( \mathbf{x}_t )) &= 
\frac{\mathbf{x}_t - \sqrt{1-\alpha_t} \bm{\epsilon}^\theta_t(\mathbf{x}_t) }{\sqrt{\alpha_t}} \label{eq:P_declaration}\\
%\qquad 
\intertext{and}
%\qquad 
\mathbf{D}_t(\bm{\epsilon}^\theta_t(  \mathbf{x}_t ))  &= \sqrt{1-\alpha_{t-1} - \sigma_t^2}
\bm{\epsilon}^\theta_t(  \mathbf{x}_t )
\label{eq:P_D_declaration}
\end{align}
are the predicted $\mathbf{x}_0$ and the direction pointing to $\mathbf{x}_t$ at timestep $t$, respectively.
The variance $\sigma_t$ is taken to be
\begin{equation}
\sigma_t = \eta_t  \sqrt{(1-\alpha_{t-1})/(1-\alpha_t)}\sqrt{1- \alpha_t/\alpha_{t-1}}. 
\end{equation}
The special case where $\eta_t = 0$ for all $t$ is called DDIM~\cite{song2020ddim}. In this setting the noise variance is $\sigma_t = 0$, so that the sampling process is deterministic and fully reversible \cite{ho2020denoising, Dhariwal2021dpmsbeatgans} (\emph{i.e.,}~$\mathbf{x}_T$ can be uniquely obtained from $\mathbf{x}_0$).
The case where $\eta_t = 1$ corresponds to the stochastic DDPM scheme~\cite{ho2020denoising}. 

\def\imga{0.22\textwidth}
\begin{figure}[tb]
%\centering
\hspace{0.25cm}
\begin{subfigure}[b]{0.95\linewidth}
\hspace{20pt}
{\scriptsize S1}
\hspace{40pt}
{\scriptsize S2}
\hspace{20pt}
$\substack{\text{S1} \\ (\mathbf{h}_t \text{from S2}) } $ 
\hspace{12pt}
$\substack{\text{S2} \\ (\mathbf{h}_t \text{from S1}) } $ \\
    \includegraphics[width=\imga]{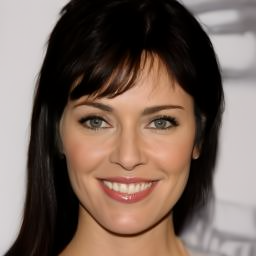}
    \includegraphics[width=\imga]{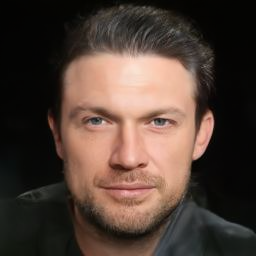}
    \includegraphics[width=\imga]{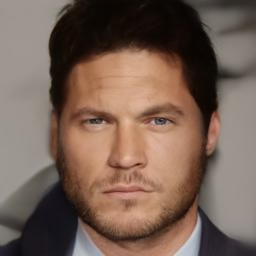}
    \includegraphics[width=\imga]{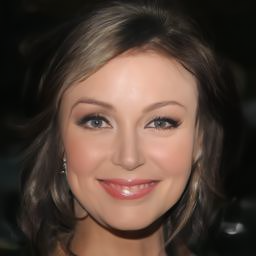}
\caption{Effect of swapping the bottleneck activation.}
\label{fig:swap_hs}    
\end{subfigure}
\\
%%%%%%%%%%%%
\begin{subfigure}[b]{0.95\linewidth} 
\mbox{}
\vspace{-2mm}
\includegraphics[width=\linewidth,trim={0 3cm 0 4cm},clip]{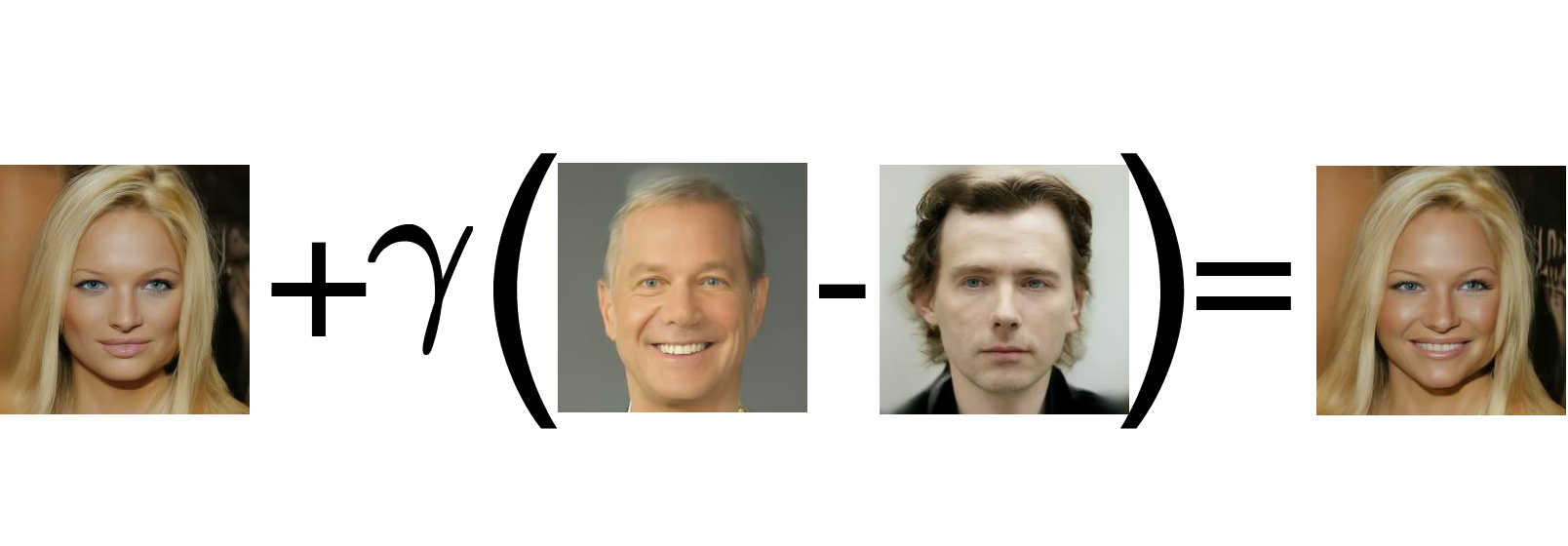}
\caption{Vector arithmetic in the semantic latent space.}
\label{fig:single_example_edit}
\end{subfigure}
\caption{
    \textbf{Illustration of properties of the $h$-space.}
    \subref{fig:swap_hs}  Swapping $\mathbf{h}_{T:1}$ between two samples, S1 and S2, swaps the semantic content without affecting background. 
    \subref{fig:single_example_edit} Adding the difference in bottleneck activation $\mathbf{h}_{T:1}$ between a smiling and non-smiling person results in a smile in a new sample. The result are shown with strength parameter $\gamma=1/5$.
}
\end{figure}

Following Kwon \etal~\cite{Kwon2022ddmhavesemantic}, we study the semantic latent space of DDMs corresponding to the activation of the bottleneck feature maps of the U-Net. We denote the concatenation of the bottleneck activation across all timesteps as $\mathbf{h}_{T:1}$ see supplementary material (SM) Sec.~\ref{sm:hspace-diagram} for illustration and additional details.
%% Definition of the semantic latent space
In \cite{Kwon2022ddmhavesemantic} image editing was performed 
via an asymetric reverse process (Asyrp), where~$\Delta\mathbf{h}_t$ is only injected into $\mathbf{P}_t$ of \eqref{eq:ddim-reverse} and not to $\mathbf{D}_t$.
Empirically, we find that Asyrp amplifies the effect of the edits but semantic editing is also possible without using Asyrp.
In this paper, we inject $\Delta \mathbf{h}_t$ into both terms of~\eqref{eq:ddim-reverse}. 
This has the benefit of only requiring a single forward pass of the U-Net at each step of the sampling process, as opposed to the two forward passes needed in Asyrp (one for $\mathbf{P}_t$ with injection and one for $\mathbf{D}_t$ without the injection).
In SM Sec.~\ref{SM:asyrp} we provide a comparison of the effect of editing with and without using Asyrp.

%% The h-space is not a complete latent representation.
The bottleneck activation $\mathbf{h}_t$ is determined directly from~$\mathbf{x}_t$ in each step of the generative process. 
It is worth noting that although most of the high-level semantic content of the generated image is determined by~$\mathbf{h}_{T:1}$, it is not a complete latent representation in the sense that it does not completely specify the generated image. We illustrate this point in Fig.~\ref{fig:swap_hs} where we swap~$\mathbf{h}_{T:1}$ between two samples while keeping~$\{\mathbf{x}_T, \mathbf{z}_{T:1}\}$ fixed. We observe that swapping~$\mathbf{h}_{T:1}$ results in a swap of the high-level semantics, like the gender, but not the background. 

%% Vector arithmetic proporty semantic latent space
A key property of $h$-space is that it obeys vector arithmetic properties which have previously been demonstrated for GANs by Radford \etal~\cite{radford2016dcgan}. Specifically, image editing can be done in $h$-space as follows. Suppose we have found a direction $\mathbf{v}_{T:1}$ associated with some semantic content that we wish to apply to a sample with latent code $\mathbf{h}_{T:1}$. Then $\mathbf{h}_{T:1}^{(\text{edit})} = \mathbf{h}_{T:1} + \gamma \mathbf{v}_{T:1}$ is the latent code of the edited image, where $\gamma$ controls the strength of the edit. In Fig.~\ref{fig:single_example_edit} we illustrate the vector arithmetic property of $h$-space by adding a difference vector which has the semantic effect of adding a smile.

% $h$-space facilitates semantic editing in DDMs the representation 
% is not complete latent representation in the sense that it . 
% Although the high level semantic content in determined by $\mathbf{h}_\mathrm{all}$ is is not a complete latent representation. 
% howfter the majority of the semantic content is present in $\mathbf{h}_t$
% is determined in a deterministic way from $\mathbf{x}_t$. 
% For example one cannot interpolate between two samples in $h$-space. 

% For example subtracting the bottleneck activations $\mathbf{h}_\mathrm{all}$ for a smiling person from a non-smiling person and adding the result to a non-smiling woman across results in the appearance of a smile with only minor changes to other attributes. 

\section{Unsupervised semantic directions}\label{sec:Unsupervised methods}

\subsection{Principal component analysis}
Our first goal is to uncover interesting semantic directions in an unsupervised fashion.
To this end, we first explore the use of principal component analysis (PCA) in $h$-space.
In the context of GANs \cite{harkonen2020ganspace}, it was shown that the principal components of a collection of randomly sampled latent codes result in semantically interpretable editing directions. 
Here we demonstrate that the same is true for DDMs if the PCA is performed in the semantic $h$-space.
Specifically, we consider PCA where we generate $n$ random samples and save the bottleneck activation $\mathbf{h}^{(i)}_t$ for each sample $i$ at all timesteps. 
Then, for each timestep $t$ we vectorize $\{\mathbf{h}^{(i)}_t\}_{i=1}^n$ and calculate the principal components. 
We use Incremental PCA \cite{ross2008incremental} in order to calculate PCA on more samples than would otherwise fit in memory. 
We define the editing direction $\mathbf{v}_{j}$ as a concatenation of the $j$'th principal component from all timesteps.
To demonstrate our method, we use Diffusers~\cite{von-platen-etal-2022-diffusers} and a DDPM\footnote{\url{https://huggingface.co/google/ddpm-ema-celebahq-256}} trained on the CelebA \cite{liu2015celeba} data set. 
Unless stated otherwise, all results use $\eta_t = 1$ during the synthesis process. 

\begin{figure}[tb]
    \centering
    \includegraphics[width=\linewidth]{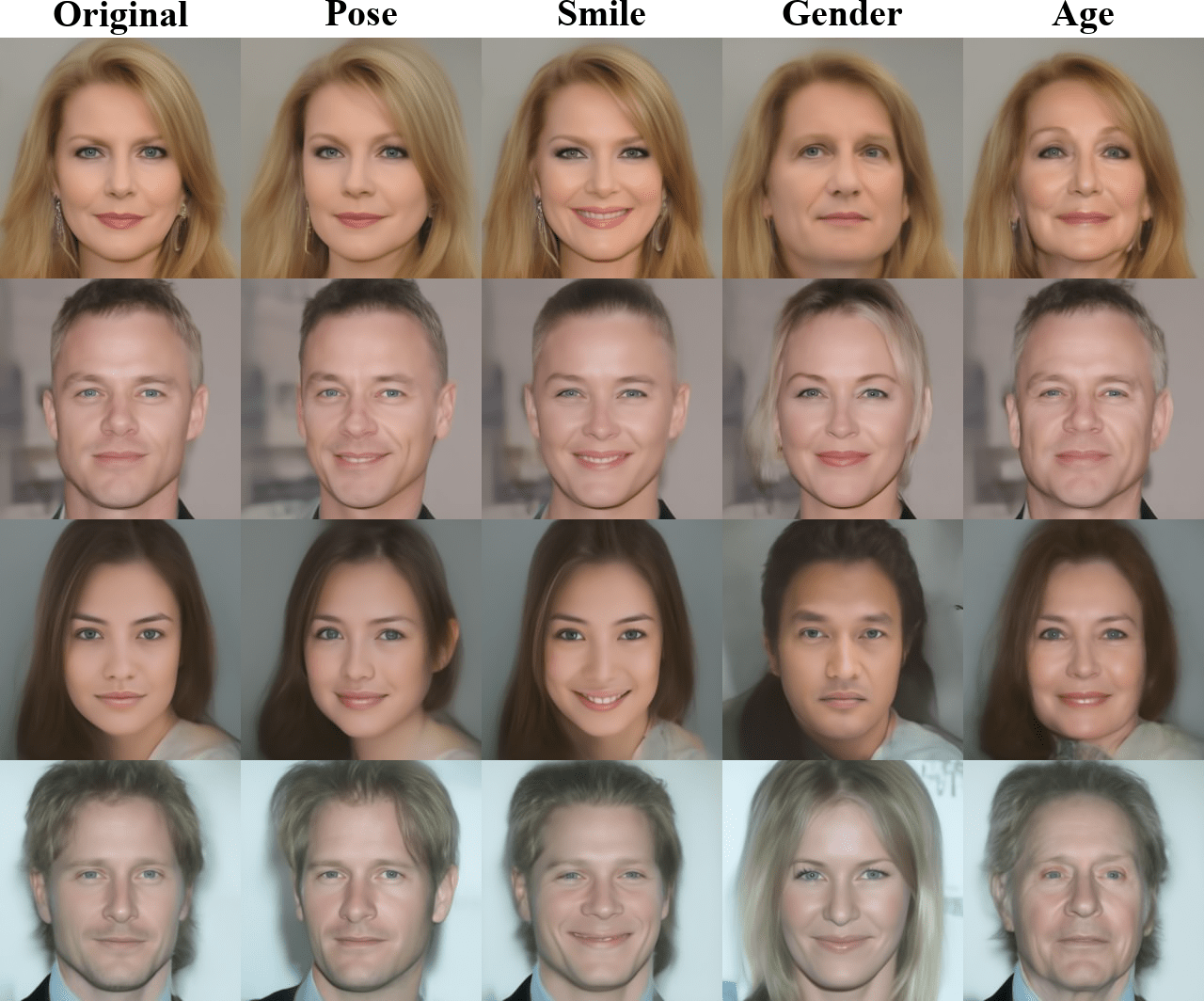}
    \caption{
    \textbf{PCA in the semantic latent space.}
    PCA in $h$-space provides a way for discovering disentangled and semantically meaningful directions. Here we show a selection of semantic edits corresponding to pose, smile, gender and age. 
    }
    \label{fig:pca}
\end{figure}

\begin{figure}[tb]
\centering
    \begin{subfigure}[b]{0.9\linewidth}
    \includegraphics[width=\linewidth]{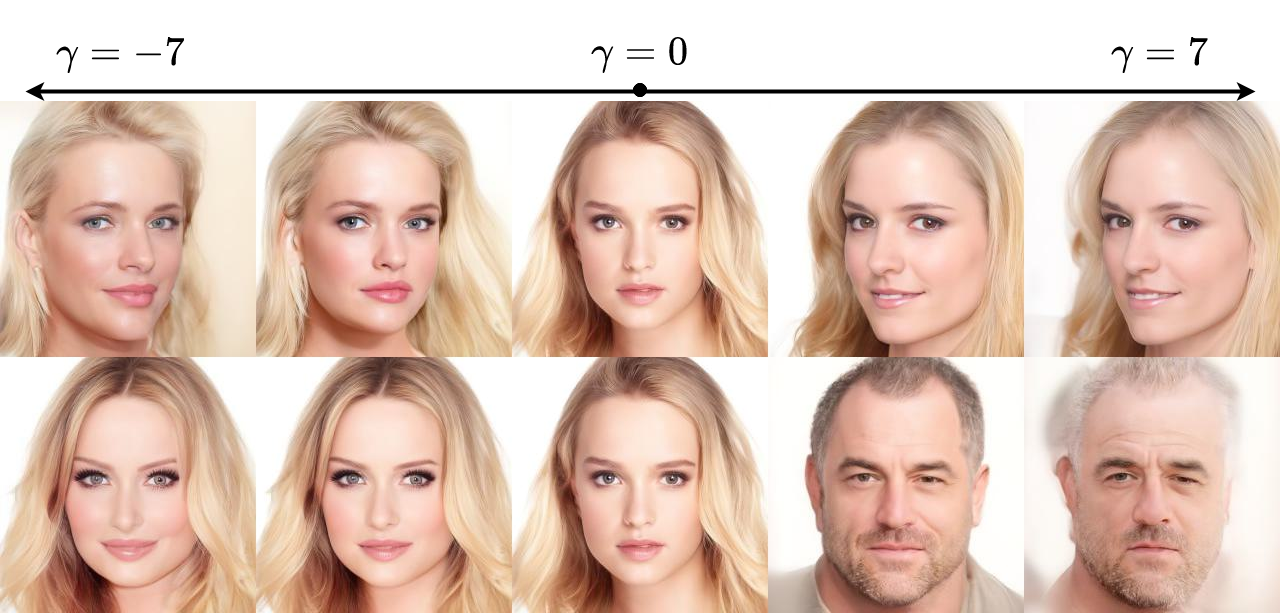}
    \caption{Two dominant PCA directions}
    \label{subfig:pca-top2}
    \end{subfigure}
    
    \begin{subfigure}[b]{0.9\linewidth}
    \includegraphics[width=\linewidth]{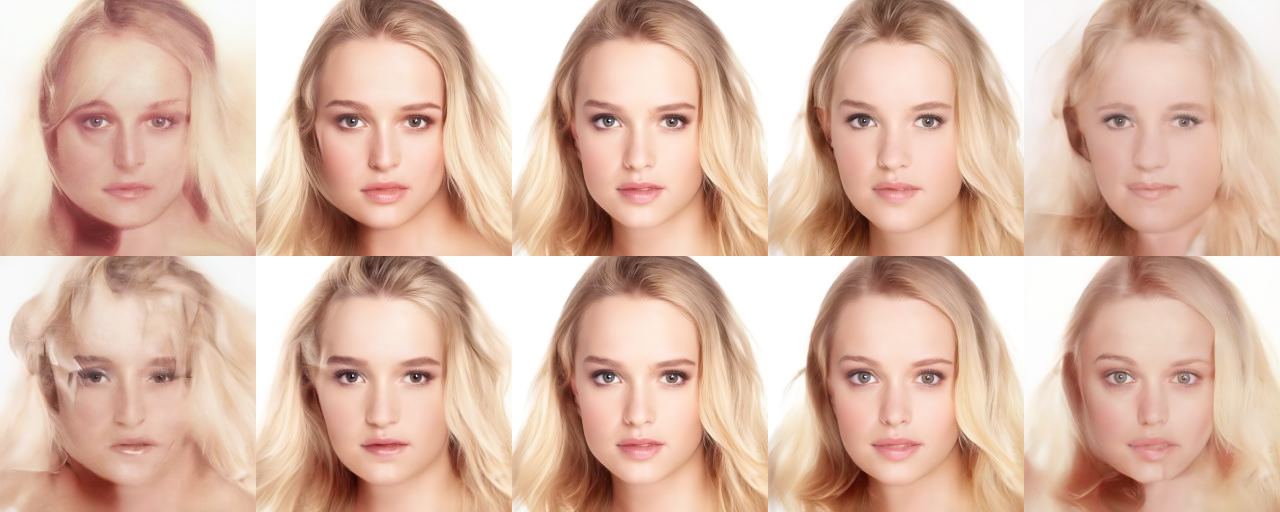}
    \caption{Random directions}
    \label{subfig:pca-random}
    \end{subfigure}

    \caption{\textbf{PCA v. random directions}
    While directions found with PCA have a clear semantic meaning, like pose and gender, interpolating along random directions results in only minor changes to the image when using the same scale. Increasing the scale results in a degradation of the image.}
\end{figure}

%The results are displayed in Fig.~\ref{fig:pca_whole}.
It can be seen that many principal directions have clear semantic interpretations, Fig.~\ref{fig:pca} demonstrates the effect of several of these directions, including directions corresponding to gender, pose, age, and smile. 
Fig.~\ref{subfig:pca-top2} and \ref{subfig:pca-random} compares the effect of applying the two dominant principal components to random directions. For a fair comparison, we set the norm of~$\Delta \mathbf{h}_t$ for the random directions to match that of the principal components. While interpolating along principal directions leads to semantically interpretable edits, shifting along random directions only induces minor changes to the image at small scales and rapid degradation of the image at larger scales. 


% \begin{figure}[tb]
% \centering
% \begin{subfigure}[b]{0.44\linewidth}
%     \begin{subfigure}[b]{\linewidth}
%     \includegraphics[width=\linewidth]{figs/pca/top2pca_annotated.png}
%     \caption{Two dominant PCA directions}
%     \label{subfig:pca-top2}
%     \end{subfigure}
    
%     \begin{subfigure}[b]{\linewidth}
%     \includegraphics[width=\linewidth]{figs/pca/random_direction.png}
%     \caption{Random directions}
%     \label{subfig:pca-random}
%     \end{subfigure}
% \end{subfigure}
% \begin{subfigure}[b]{0.5\linewidth}
%     \includegraphics[width=\linewidth]{figs/pca/PCA-Plot-new.png}
%     \caption{Selection of semantic directions unveiled by PCA.}
%     \label{fig:pca}
% \end{subfigure}
% \caption{%
% \textbf{PCA in the semantic latent space.}
% \subref{subfig:pca-top2}-\subref{subfig:pca-random} While directions found with PCA have a clear semantic meaning, like pose and gender, interpolating along random directions results in only minor changes to the image when using the same scale. Increasing the scale results in a degradation of the image.
% \subref{fig:pca} PCA in $h$-space provides a way for discovering disentangled and semantically meaningful directions. Here we show edits corresponding to pose, smile, gender and age. 
% }
% \label{fig:pca_whole}
% \end{figure}

\subsection{Discovering image-specific semantic edits}
\begin{figure*}[t]
\centering
\includegraphics[width=0.98\linewidth]{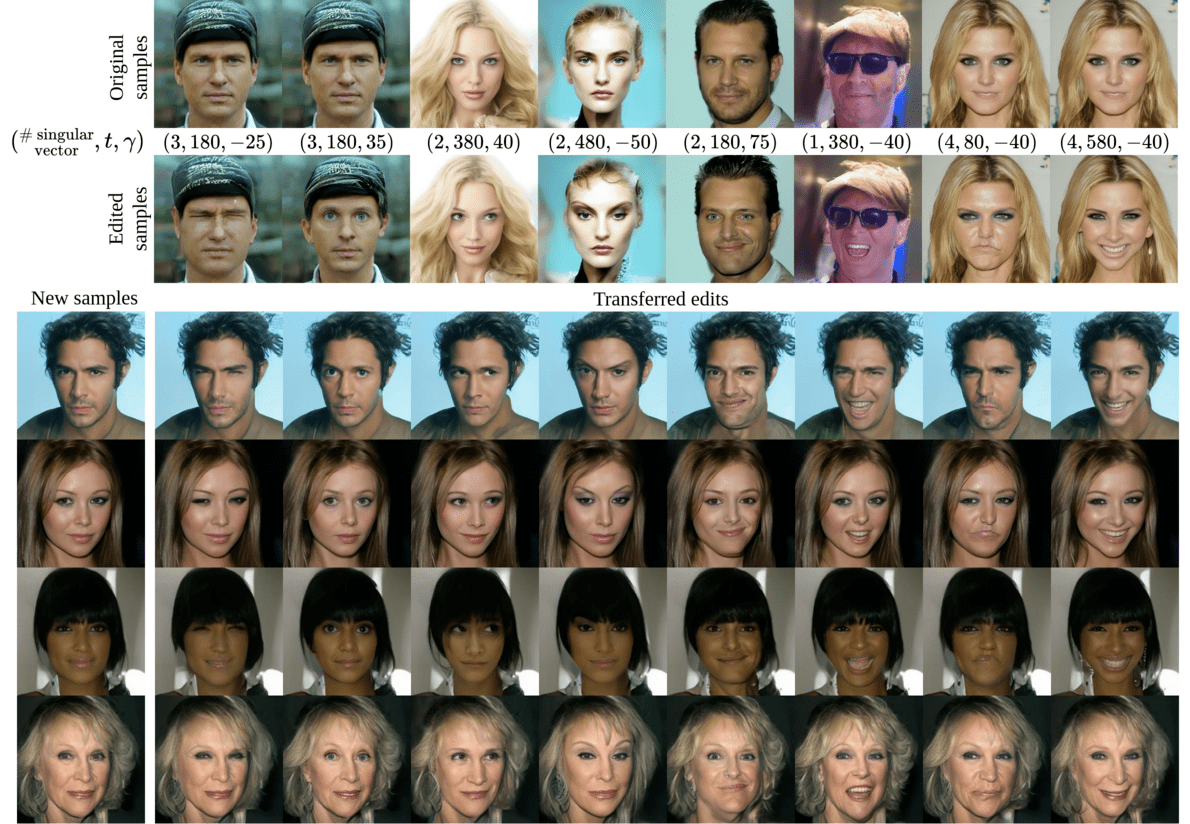}
\caption{
\textbf{Unsupervised image-specific edits.}
Spectral analysis of the Jacobian of $\bm{\epsilon}_t^\theta$ yields directions corresponding to localized changes in the generated image, \eg eyes opening/closing and raising of the eyebrows.
Although this method is image-specific, directions found for one sample can be transferred to others, where they result in semantically similar edits. }
\label{fig:poweriter}
\end{figure*}

%Motivation. 
% Global directions have the advantage that they are computed based on many samples and are thus applicable to any image. However, many desirable semantic edits are not applicable to all images. 
% For example, a direction that opens or closes eyes is clearly irrelevant for a person wearing dark sunglasses. Therefore, we now focus on finding image-specific semantic directions.

The directions found with PCA are computed based on many samples and tend to find global changes such as pose and gender, while more local changes like the closing of the eyes are absent. The smile direction is the only direction we observed where the semantic changes are localized to a specific region like the mouth. In the following, we present a method to find directions that are specific to a single image and region of interest. 

% What we do - approach.  
To find directions specific to a single image we wish to find a set of orthogonal directions in $h$-space that induce the largest change in the prediction of the clean image $\mathbf{P}_t(\bm{\epsilon}^\theta_t( \mathbf{x}_t ))$ at every timestep. 
This is equivalent to finding the directions that change $\bm{\epsilon}^\theta_t( \mathbf{x}_t )$ the most (see SM Sec.~\ref{SM:noteonnoiseprediction}).
For small perturbations, these directions are the top right-hand singular vectors of the Jacobian of $\bm{\epsilon}^\theta_t$ with respect to $\mathbf{h}_{t}$. 
Due to the skip-connections in the U-Net, the output of the network depends on both $\mathbf{x}_{t}$ and $\mathbf{h}_{t}$. Yet, here we only consider the dependency on the latent variable $\mathbf{h}_{t}$.
In the following, we  denote the Jacobian of $\bm{\epsilon}^\theta_t$ by $\mathbf{J}_t$ and  its singular value decomposition (SVD) as
\begin{equation}
\mathbf{J}_t \triangleq \frac{\partial \bm{\epsilon}^\theta_t(\mathbf{x}_t, \mathbf{h}_t)}{\partial\mathbf{h}_t} 
= \mathbf{U}_t\bm{\Sigma}_t \mathbf{V}^\mathrm{T}_t. 
\end{equation}

The right singular vectors corresponding to the largest singular values, (the columns of $\mathbf{V}_t$) are the set of orthogonal vectors in $h$-space which perturb the predicted image the most. 
Note that for each timestep $t$, we have a different set of directions.  
In practice, we find that semantically interesting effects are obtained by applying directions found at timestep $t$ across all timesteps. Thus, computing~$k$ directions per timestep provide us $kT$ potential edits in each of the~$T$ timesteps. 
In SM Sec.~\ref{SM:jacobiantimesteps}, we illustrate the qualitative difference between directions computed at different timesteps.  

% The computation trick
In practice, calculating $\mathbf{J}_t$ directly is computationally expensive. Instead, we find the dominant singular vectors by power-iteration over the matrix $\mathbf{J}_t^\mathrm{T} \mathbf{J}_t$, whose eigenvectors are precisely the right singular vectors of $\mathbf{J}_t$. 
Each iteration requires multiplication by $\mathbf{J}_t^\mathrm{T} \mathbf{J}_t$, which can be computed without ever storing the Jacobian matrix in memory. 
Specifically, for any vector $\mathbf{v}$, the  product $\mathbf{J}_t^\mathrm{T}\mathbf{J}_t\mathbf{v}$ can be computed as 
\begin{align}
\label{eq:poweriter_trick}
    \mathbf{J}_t^\mathrm{T}\mathbf{J}_t\mathbf{v} &= \frac{\partial}{\partial\mathbf{h}_t}
    \left\langle \bm{\epsilon}^\theta_t(\mathbf{x}_t,\mathbf{h}_t) ,\mathbf{J}_t\mathbf{v} \right \rangle\\
%\qquad
\intertext{with}
%\qquad
    \mathbf{J}_t\mathbf{v} &= \left.\frac{\partial}{\partial a} \bm{\epsilon}^\theta_t(\mathbf{x}_t,\mathbf{h}_t  + a \mathbf{v})\right|_{a=0}.
\end{align}

\begin{algorithm}[t]
\caption{Jacobian subspace iteration}\label{alg:cap}
\begin{algorithmic}
\Require $ \mathbf{f} : \mathbb{R}^{d_{\text{in}}} \to \mathbb{R}^{d_{\text{out}}} $, $ \mathbf{h} \in  \mathbb{R}^{d_{\text{in}}} $ and $ \mathbf{V} \in  \mathbb{R}^{d_{\text{in}} \times k} $ 
\Ensure $ (\mathbf{U}, \mathbf{\Sigma}, \mathbf{V}^\mathrm{T}) $ -- $k$ largest singular values and singular vectors of the Jacobian $ {\partial \mathbf{f} }/{ \partial \mathbf{h}}$
\State $\mathbf{y} \gets \mathbf{f}(\mathbf{h})$ % , \;  a \gets 1  $
\If{$\mathbf{V}$ is empty}
    \State $\mathbf{V} \gets $ i.i.d.\@ standard Gaussian samples
\EndIf
\State $  \mathbf{Q},\mathbf{R} \gets \mathrm{QR}(\mathbf{V}) $
\Comment{Reduced QR decomposition}
\State $\mathbf{V} \gets \mathbf{Q}$
\Comment{Ensures $ \mathbf{V}^\mathrm{T} \mathbf{V} = \mathbf{I} $}
\While{stopping criteria}
% \State $\mathbf{B} \gets \mathbf{h}-\mathbf{U} $ 
% \Comment{$\mathbf{h}$ broadcasted}
\State $\mathbf{U} \gets \partial \mathbf{f} ( \mathbf{h} \mathbf{1}_k^\mathrm{T} +a \mathbf{V} ) / \partial a $ at $ a = 0$
\Comment{Batch forward}
\State $\hat{\mathbf{V}} \gets \partial (\mathbf{U}^\mathrm{T}\mathbf{ y })/\partial \mathbf{h}$
\State $\mathbf{V},\mathbf{\Sigma^2}, \mathbf{R} \gets \mathrm{SVD}(\hat{\mathbf{V}})$
\Comment{Reduced SVD}
\EndWhile
\State Orthonormalize $\mathbf{U}$
\end{algorithmic}
\end{algorithm}

Our algorithm is summarized in Alg.~\ref{alg:cap} and uses \eqref{eq:poweriter_trick} to calculate the singular vectors of the Jacobian of an arbitrary vector-valued function $\mathbf{f}$. 
The algorithm starts by randomly initializing a set of vectors $\{\mathbf{v}_i \}_{i=1}^k$ and iterative computes~\eqref{eq:poweriter_trick} using automatic differentiation while enforcing orthogonality among the singular vectors. 
Importantly, it was shown that batched power iteration with an orthogonalization step, such as presented here, is guaranteed to converge to the SVD of positive semi-definite matrices \cite[Ch.~5]{saad2011numerical}. 

% Few words are in place regarding implementation.
Regarding implementation, in \eqref{eq:poweriter_trick} we compute a derivative of high dimensional output w.r.t.\@ a scalar. 
This is efficiently done by utilizing forward mode automatic differentiation.
Further, \eqref{eq:poweriter_trick} can be calculated in parallel for multiple vectors using the batched Jacobian-vector product, \eg in Pytorch. 
Since, parallel calculation of a large number of vectors can be memory intensive, we give a sequential variant of Alg.\ref{alg:cap} in SM, Sec.~\ref{SM:jacobian}.

%Our proposed method successfully identifies semantically meaningful directions that correspond to highly localized semantic changes in the image, e.g. closing or opening of the eyes and mouth, or raising of the eyebrows. We show a selection of such localized edits at the top of Fig.~\ref{fig:poweriter}.
Our method identifies semantically meaningful directions for localized semantic image changes (e.g., eye and mouth movements), as shown in Fig.~\ref{fig:poweriter}. Although these directions are image-specific, they consistently produce similar changes across different images, demonstrating the effectiveness and generalizability of our approach.
%
%While the semantic directions found by this method are image-specific and may vary depending on the sample analyzed, we find that they result in the same localized changes when applied across different images. 
This is illustrated in the lower part of Fig.~\ref{fig:poweriter} where each of the found editing directions is applied with the same magnitude $\gamma$ across a selection of samples. These results suggest that our approach is effective in identifying meaningful semantic directions that generalize across different images.

\begin{figure*}[h!]%[htb]
    \centering
    \begin{subfigure}[b]{0.49\linewidth}
        \includegraphics[width=\linewidth]{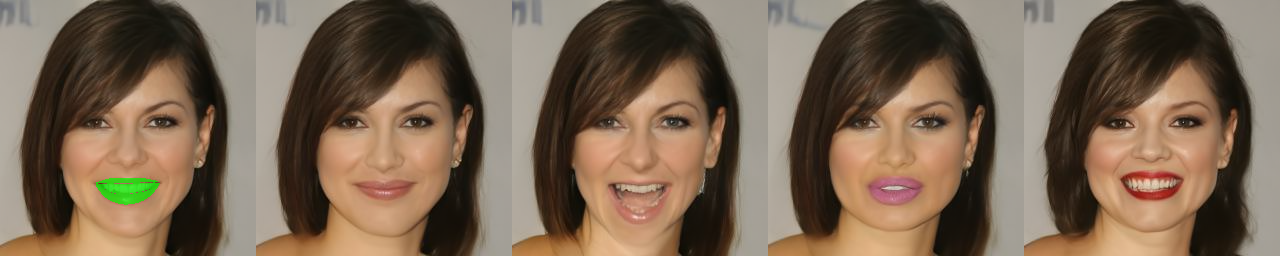}
        \includegraphics[width=\linewidth]{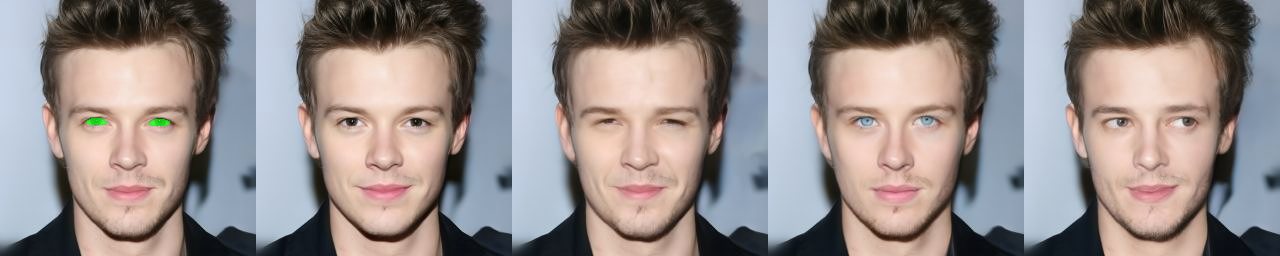}
    \end{subfigure}
    \begin{subfigure}[b]{0.49\linewidth}
        \includegraphics[width=\linewidth]{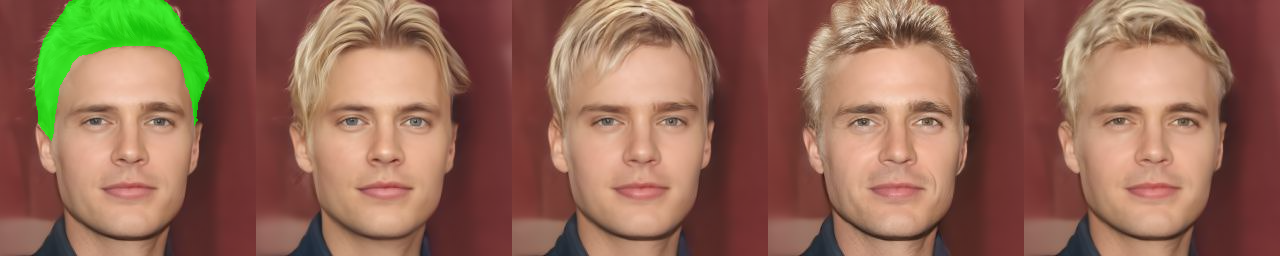}
        \includegraphics[width=\linewidth]{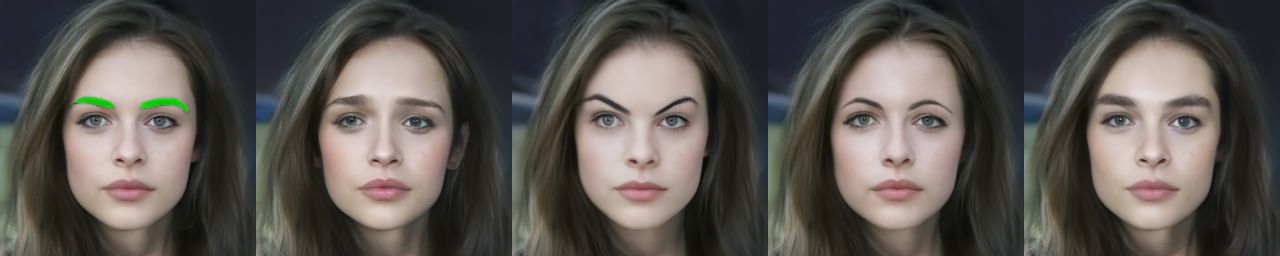}
    \end{subfigure}
    \caption{
    \textbf{Region-specific edits.}
        Given a mask specifying a region of interest, our method can be guided to focus on finding directions which change only the target area. The first column shows the input with the mask shown in green. 
    }
    \label{fig:regionspecific}
\end{figure*}
%\FloatBarrier

If additional information is available in the form of a mask specifying a region of interest, our method can be naturally extended by applying the mask to the noise prediction $\widetilde{\bm{\epsilon}}^\theta_t$ in order to find directions in $h$-space that change a specific region the most rather than the whole image. We seek the singular vectors of the Jacobian of the masked output of the U-net. We define the a masked Jacobian $\mathbf{J}_t^\text{masked}$ as 
\begin{align}
\mathbf{J}_t^\text{masked} &= \partial  \widetilde{\bm{\epsilon}}^\theta_t(\mathbf{x}_t,\mathbf{h}_t) / \partial \mathbf{h}_t, \\
%\quad \quad 
\widetilde{\bm{\epsilon}}^\theta_t(\mathbf{x}_t,\mathbf{h}_t) &= \bm{\epsilon}^\theta_t(\mathbf{x}_t,\mathbf{h}_t)\odot \mathbf{M},    
\end{align}
where $\odot$ denoted the Hadamard product and $\mathbf{M}$ is a binary mask corresponding to a region of interest. We show examples of such region-specific edits in Fig.~\ref{fig:regionspecific}.

\section{Supervised discovery of semantic directions}\label{sec:Supervised methods}
%%% Motivation 
%%% The unsupervised method are great but require manual inspection.
While the methods we presented in Sec.~\ref{sec:Unsupervised methods} discover interpretable semantic directions in a fully unsupervised fashion, their effects must be interpreted manually. In this section, we demonstrate a simple supervised approach to obtain latent directions corresponding to well-defined labels.

\paragraph{Linear semantic directions from examples}
The vector arithmetic property of $h$-space suggests an intuitive method for discovering semantically meaningful directions, by providing positive and negative examples of a desired attribute. 
Let $\{(\mathbf{x}_i^-,\mathbf{x}_i^+ )\}_{i=1}^n$ be a collection of generated images, such that all $\mathbf{x}_i^+$ have a desired attribute that is absent in~$\mathbf{x}_i^-$, \eg a smile, old age, glasses, \etc. 
Let~$\mathbf{q}_i^-$ and $\mathbf{q}_i^+$ denote the latent representation corresponding to the images~$\mathbf{x}_i^-$ and~$\mathbf{x}_i^+$. Then, we can find a semantic direction~$\mathbf{v}$ as 
\begin{equation}\label{eq:linear_direction}
    \mathbf{v} = \frac{1}{n} \sum_{i=1}^n\left(\mathbf{q}_i^+-\mathbf{q}_i^-\right).
\end{equation}

\begin{figure}[tb]
\centering
    \begin{subfigure}[b]{0.99\linewidth}
        \includegraphics[width=\linewidth]{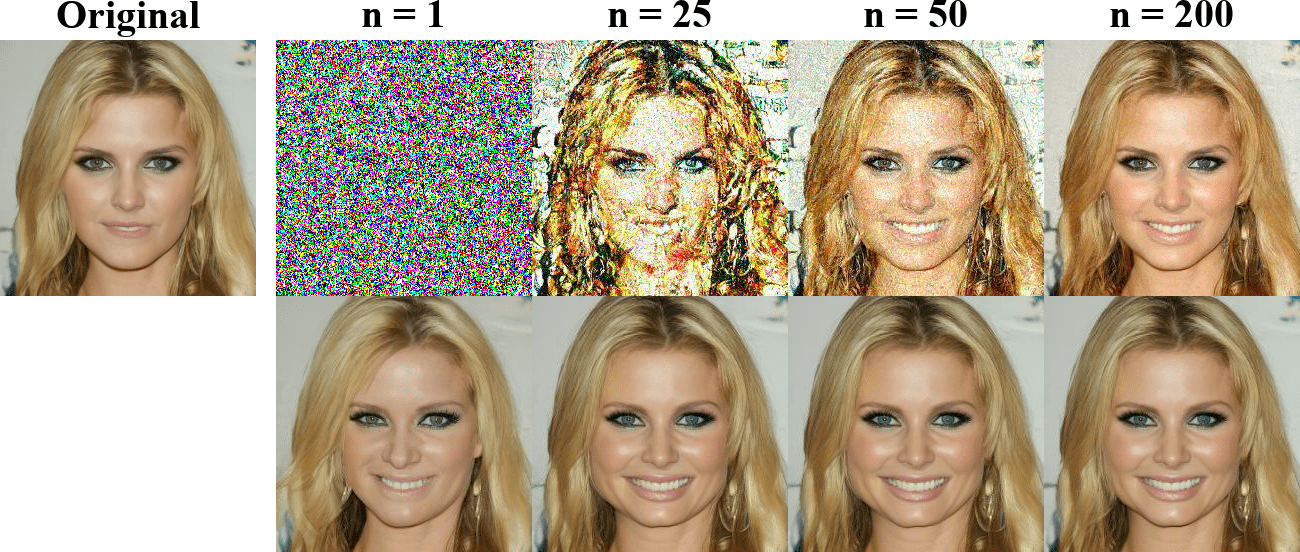}
        \caption{Editing in $h$-space vs.\@ using $\mathbf{x}_T$.}
        \label{fig:h_vs_ddim_numsamples}
    \end{subfigure}\\[2mm]
    \begin{subfigure}[b]{0.99\linewidth}
        \includegraphics[width=\linewidth]{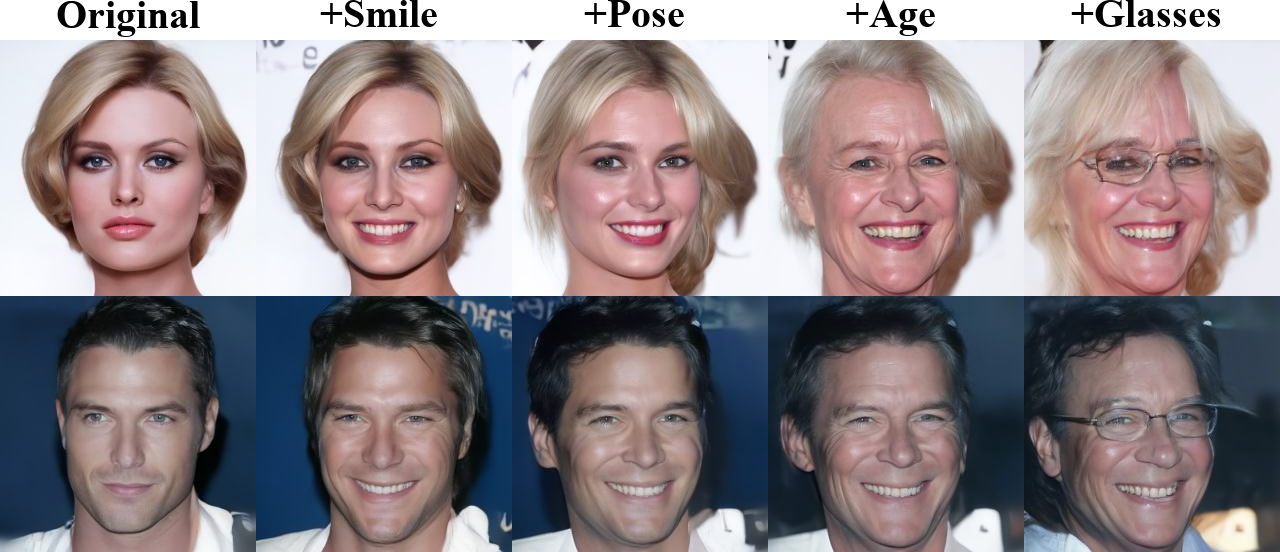}
        \caption{Sequential manipulation.}
        \label{fig:anycost_sequential}
    \end{subfigure}
    \caption{
    \textbf{Editing properties of $h$-space.}
    \subref{fig:h_vs_ddim_numsamples} 
    A qualitative comparison of the editing effect using $\mathbf{x}_T$ (top) and $\mathbf{h}_{T:1}$ (bottom). Latent variables using a smiling direction found by \eqref{eq:linear_direction}.
    While the direction in $h$-space converges with a few labeled examples, more than $200$ are required to achieve a similar result using $\mathbf{x}_T$ as the latent variable.
    \subref{fig:anycost_sequential} 
    Directions found with our method can be combined with one another. Here, we sequentially accumulate four effects, starting from a single effect in the 2nd column up to four effects in the 5th column.
    }
\end{figure}

% Effect of editing in the DDIM Noise space vs $h$-space
Note that this method can be applied using either $\mathbf{h}_{T:1}$ or $\mathbf{x}_T$ as the latent variable. 
However, defining semantic directions using $\mathbf{h}_{T:1}$ as the latent variable requires far fewer samples than using $\mathbf{x}_T$. Figure~\ref{fig:h_vs_ddim_numsamples} illustrates this for DDIM ($\eta_t = 0$) for a direction corresponding to smile where \eqref{eq:linear_direction} is calculated using a varying number of samples.

\paragraph{Classifier annotation}\label{sec:supervised}
We now propose to find linear semantic directions by using pretrained attribute classifiers to annotate samples generated by the model. Using the attribute classifier from \cite{lin2021anycost}, we annotate samples with probabilities corresponding to the $40$ classes from CelebA \cite{liu2015celeba}, and use Hopenet \cite{Ruiz2018hopenet} to predict pose (yaw, pitch, and roll).
We sort the annotated samples according to the attribute scores and select the samples with the highest and lowest scores from each class as the positive and negative examples respectively. 
We then calculate semantic directions corresponding to the different attributes using the method given in  \eqref{eq:linear_direction}.

\begin{figure}[t]
    \centering
\begin{subfigure}[b]{0.48\linewidth}
\includegraphics[width=\textwidth]{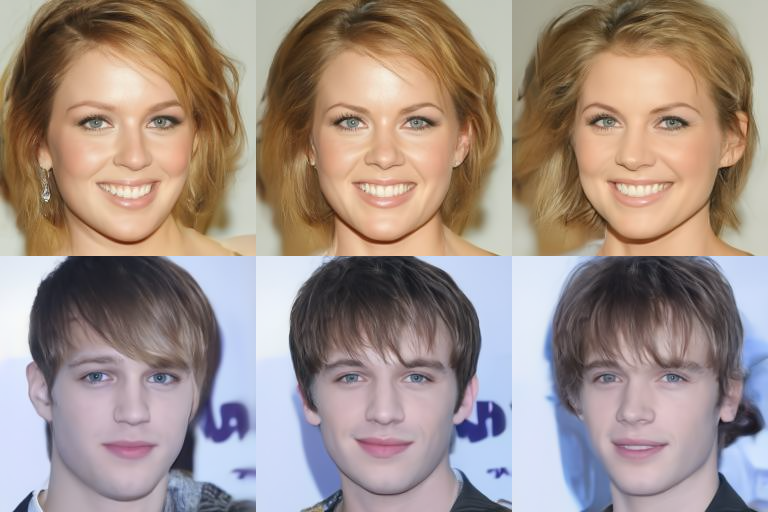}
\caption{Yaw}
\end{subfigure}
\begin{subfigure}[b]{0.48\linewidth}
\includegraphics[width=\textwidth]{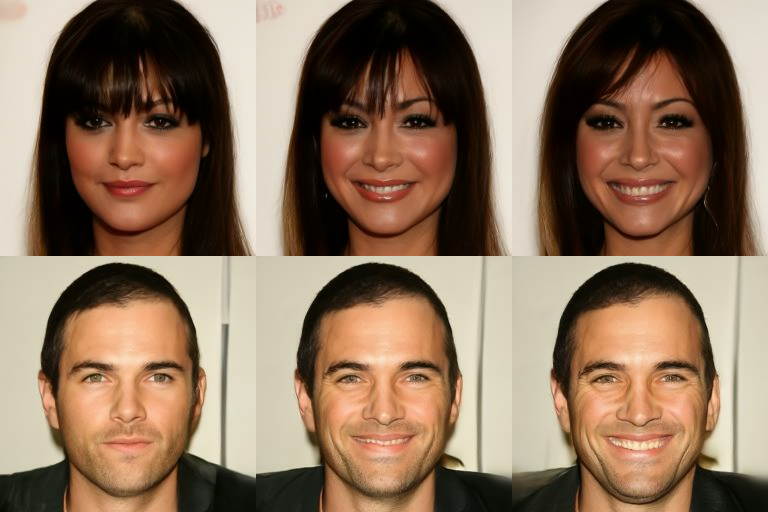}
\caption{Smile}
\end{subfigure}
           
\begin{subfigure}[b]{0.49\linewidth}
\includegraphics[width=\textwidth]{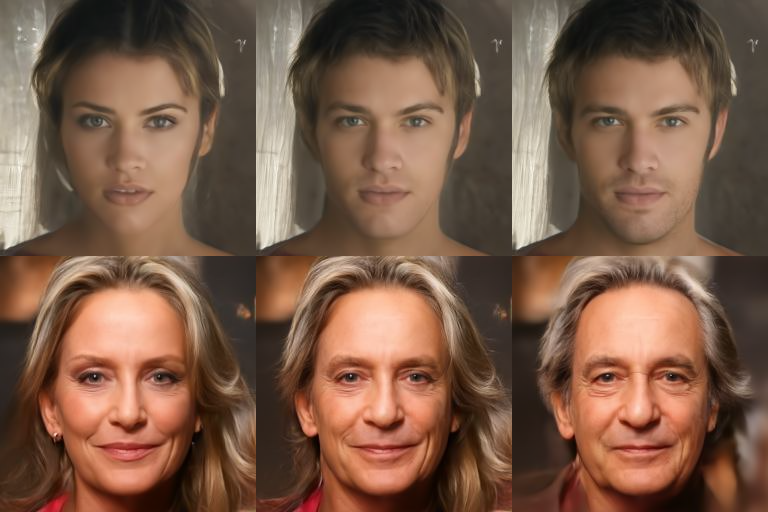}
\caption{Gender}
\end{subfigure}
\begin{subfigure}[b]{0.49\linewidth}
\includegraphics[width=\textwidth]{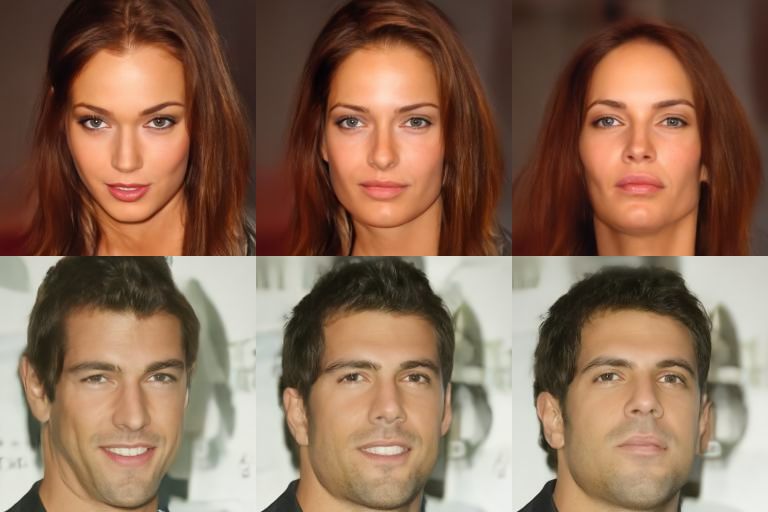}
\caption{Pitch}
\end{subfigure}
            
\begin{subfigure}[b]{0.49\linewidth}
\includegraphics[width=\textwidth]{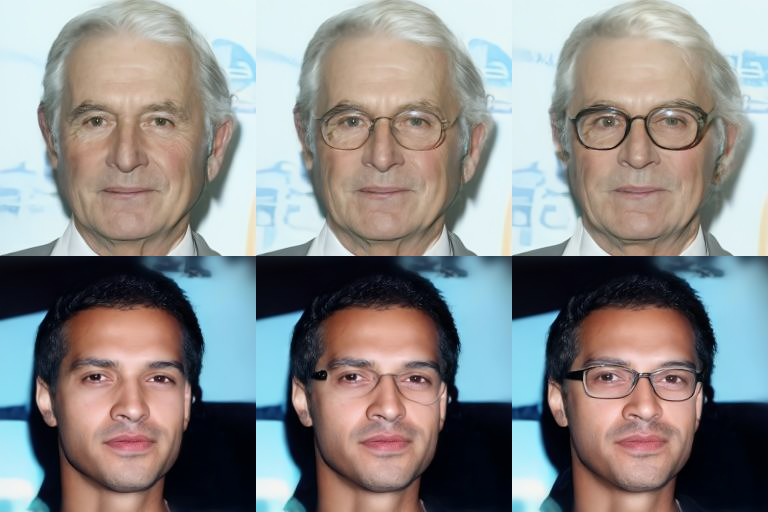}
\caption{Glasses}
\end{subfigure}
\begin{subfigure}[b]{0.49\linewidth}
\includegraphics[width=\textwidth]{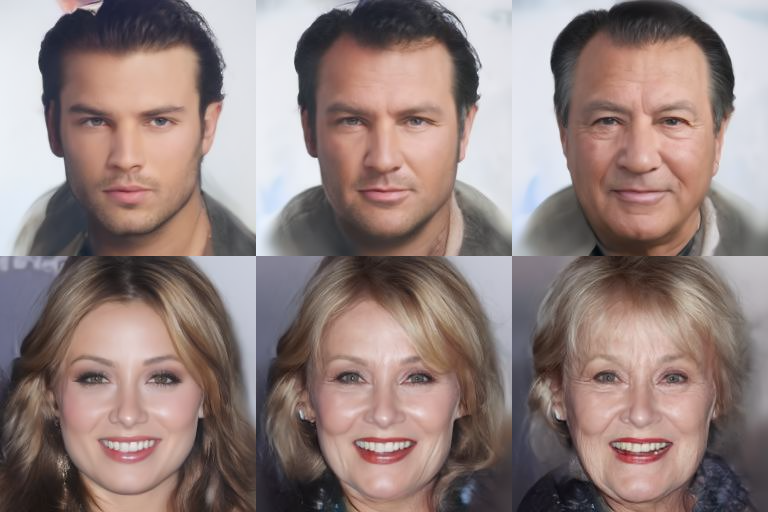}
\caption{Age}
\end{subfigure}

\caption{\textbf{Single attribute manipulation.}
Using a domain-specific binary attribute classifier, we find linear directions in $h$-space corresponding to a variety of semantic edits.}
\label{fig:single_attr}
\end{figure}

%% Attribute manipulation 
As shown in Fig.~\ref{fig:single_attr}, we can successfully find semantic directions controlling a wide selection of meaningful attributes like yaw, smile, gender, glasses, and age. 
%% Sequential Manipulation
Furthermore, directions calculated by \eqref{eq:linear_direction} can be applied in combination with one another. 
For example, adding~$\Delta \mathbf{h}_{T:1}$ for two attributes, like pose and smile, results in an image where both attributes are changed. 
Fig.~\ref{fig:anycost_sequential} illustrates sequential editing, showcasing changes in expression followed by pose, age, and eyeglasses for two samples. 
In SM Sec.~\ref{SM:bu3dfe} we show that this method can be applied to find directions corresponding to facial expressions using DDIM inversion and a real facial expression dataset \cite{yin2006bu3dfe} as supervision.

% \begin{figure*}[tb]
%     \centering
%         \begin{subfigure}[b]{0.19\linewidth}
%             \includegraphics[width=\textwidth]{figs/anycost/yaw2.png}
%             \caption{Yaw}
%             \end{subfigure}
%             \begin{subfigure}[b]{0.19\linewidth}
%             \includegraphics[width=\textwidth]{figs/anycost/smiling2.png}
%             \caption{Smile}
%             \end{subfigure}
%             \begin{subfigure}[b]{0.19\linewidth}
%             \includegraphics[width=\textwidth]{figs/anycost/gender2.png}
%             \caption{Gender}
%             \end{subfigure}
%             %\\
%             % \begin{subfigure}[b]{0.15\linewidth}
%             % \includegraphics[width=\textwidth]{figs/anycost/pitch2.png}
%             % \caption{Pitch}
%             % \end{subfigure}
%             \begin{subfigure}[b]{0.19\linewidth}
%             \includegraphics[width=\textwidth]{figs/anycost/glasses2.png}
%             \caption{Glasses}
%             \end{subfigure}
%             \begin{subfigure}[b]{0.19\linewidth}
%             \includegraphics[width=\textwidth]{figs/anycost/age2.png}
%             \caption{Age}
%         \end{subfigure}
%         \caption{\textbf{Single attribute manipulation.}
%         Using a domain-specific binary attribute classifier, we find linear directions in $h$-space corresponding to a variety of semantic edits.}
%         \label{fig:single_attr}
% \end{figure*}

\begin{figure*}[tb]
    \centering
    \includegraphics[width=\linewidth]{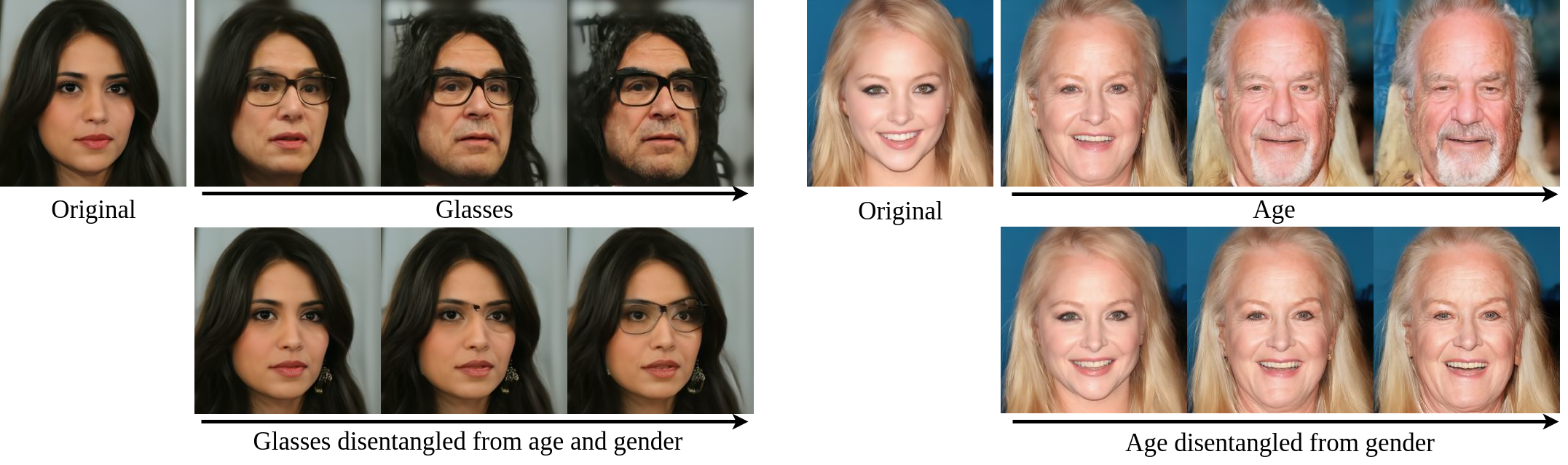}
    \caption{\textbf{Disentanglement of semantic directions.}
    Given a direction that is entangled with other attributes, we can create a disentangled direction by removing the projection onto undesired semantics. The top row shows the original direction, whereas the bottom row shows the disentangled direction.}
    \label{fig:conditional_anycost}
\end{figure*}

\paragraph{Disentanglement of semantic directions}
Latent directions found by \eqref{eq:linear_direction} might be semantically entangled, in the sense that editing in the direction corresponding to some desired attribute might also induce a change in some other undesired attributes. For example, a direction for eyeglasses may also affect the age if it correlates with eyeglasses in the training data. 
% Conditional manipulation
To remedy this, we propose conditional manipulation in $h$-space in a way similar to what was suggested in the context of GANs by Shen \etal~\cite{Shen2020Interfacegan,Shen2020InterfaceganTPAMI}. Let $\mathbf{v}_1$ and $\mathbf{v}_2$ be two linear semantic directions, where the two corresponding semantic attributes are entangled. We can define a new direction $\mathbf{v}_{1\perp 2}$ which only affects the semantics associated with $\mathbf{v}_1$, without changing the semantics associated with $\mathbf{v}_2$. This is done simply by removing from $\mathbf{v}_1$ the projection of $\mathbf{v}_1$ onto $\mathbf{v}_2$, namely $\mathbf{v}_{1\perp 2} = \mathbf{v}_1 - \langle \mathbf{v}_1  , \mathbf{v}_2 \rangle /  \| \mathbf{v}_2\|^2   \mathbf{v}_2$.
%% Disentangeling multiple semantics
In case of conditioning on multiple semantics simultaneously, our aim is to remove the effects of a collection of $k$ directions $\{\mathbf{v}_i\}_{i=1}^k$ from a primal direction $\mathbf{v}_0$ in order to define a new direction $\mathbf{v}$ which only affects the target attribute. 
This can be done by constructing the matrix $\mathbf{V} = [\mathbf{v}_1, \mathbf{v}_2, \cdots, \mathbf{v}_k]$ and projecting $\mathbf{v}_0$ onto the orthogonal complement of the column space of $\mathbf{V}$ by
\begin{equation}\label{eq:condition_multiple_semantics}
\mathbf{v} =  \left[\mathbf{I} -  \mathbf{V}\left(\mathbf{V}^\mathrm{T}\mathbf{V}\right)^{-1} \mathbf{V}^\mathrm{T} \right] \mathbf{v}_0.
\end{equation}  
The resulting direction will be disentangled from each of the directions $\{\mathbf{v}_i\}$, meaning that moving a sample along this new direction will result in a large change in the attribute associated with $\mathbf{v}_0$ while minimally affecting the attributes associated with the other directions. 
Figure.~\ref{fig:conditional_anycost} visualizes the effect of interpolating in the directions of age and eyeglasses for two samples. As can be seen, these directions are entangled with gender and age, respectively. By using our method we can successfully remove the entanglement and define a direction which only affects age or the presence of glasses. 

%% Conditional manipulation
%In some cases, dataset biases can cause the directions to be entangled with one another. For example, we observe that the direction for ``glasses'' is entangled with ``age'', which may be explained by the fact that these attributes are correlated in the training data. 

\def\rot{0}
\setlength{\tabcolsep}{3pt}
\begin{table*}
\caption{\textbf{Evaluation of disentanglement strategy.} 
We quantitatively evaluate the effect of disentangling semantic directions using linear projection. 
The rows correspond to the applied directions, while the columns correspond to the effect of the edits according to CLIP. 
We draw and edit 100 random samples and repeat the experiment 10 times with different seeds and report the mean and standard deviations. The strongest effect in each row is highlighted.
}
\label{tab:disentanglement}
% \begin{subtable}[t]{0.48\textwidth}
\resizebox{\linewidth}{!}{%
\begin{tabular}{l|ccccc|ccccc}  
\diagbox[width=5em]{Edit}{Effect} 
&\rotatebox{\rot}{Smile}&\rotatebox{\rot}{Glasses}&\rotatebox{\rot}{Age}&\rotatebox{\rot}{Gender}&\rotatebox{\rot}{Hat}
&\rotatebox{\rot}{Smile}&\rotatebox{\rot}{Glasses}&\rotatebox{\rot}{Age}&\rotatebox{\rot}{Gender}&\rotatebox{\rot}{Hat}\\
% \midrule
\hline 
\\[-8pt]
& \multicolumn{5}{c}{Original directions}  &  \multicolumn{5}{c}{Disentangled directions}  
\\[2pt]
\hline
Smile       &0.26$\pm$0.02 & 0.29$\pm$0.02 &0.08$\pm$0.02 & \textbf{0.31$\pm$0.04} & 0.07$\pm$0.01
            & \textbf{0.24$\pm$0.02} & 0.20$\pm$0.02 &0.04$\pm$0.02 & 0.09$\pm$0.03 & 0.03$\pm$0.01\\
Glasses     &0.48$\pm$0.02 & 0.32$\pm$0.02 & \textbf{0.68$\pm$0.03} & 0.66$\pm$0.04 & 0.14$\pm$0.02
            &0.22$\pm$0.01 & \textbf{0.38$\pm$0.02} &0.13$\pm$0.02 & 0.07$\pm$0.03 & 0.36$\pm$0.02\\
Age         &0.07$\pm$0.01 & 0.40$\pm$0.03 & \textbf{0.74$\pm$0.03} & 0.66$\pm$0.04 & 0.18$\pm$0.01
            &0.02$\pm$0.02 & 0.38$\pm$0.03 &\textbf{0.59$\pm$0.04} & 0.16$\pm$0.03 & 0.04$\pm$0.02\\
Gender      &0.40$\pm$0.02 & 0.28$\pm$0.03 &0.58$\pm$0.03 & \textbf{0.66$\pm$0.04} & 0.09$\pm$0.02
            &0.20$\pm$0.02 & 0.01$\pm$0.01 &0.08$\pm$0.02 & \textbf{0.39$\pm$0.03} & 0.07$\pm$0.02\\
Hat         &0.42$\pm$0.02 & 0.39$\pm$0.02 &0.37$\pm$0.03 & \textbf{0.66$\pm$0.04} & 0.41$\pm$0.02
            &0.13$\pm$0.01 & 0.03$\pm$0.03 &0.02$\pm$0.03 & 0.02$\pm$0.09 & \textbf{0.44$\pm$0.02}\\
\hline
\end{tabular}
}
\end{table*} 

%%%%% NEW NUMBERS FOR THE TABLE
%for directions 
% ["Smiling", "Eyeglasses", "Young", "Male", "Wearing_Hat"]
% Primal directions
% [[0.2654 0.1919 0.1809 0.1771 0.164 ]
% [0.3416 0.3591 0.3997 0.2698 0.2296]
% [0.1926 0.3574 0.66   0.1687 0.1583]
% [0.397  0.3096 0.5938 0.7007 0.2108]
% [0.2401 0.2341 0.1626 0.2744 0.3093]]
 
% Directions with conditioning
% [[0.2695  0.1929  0.10443 0.2269  0.1696 ]
% [0.1768  0.3167  0.1254  0.1079  0.1954 ]
% [0.1368  0.361   0.5537  0.1346  0.1343 ]
% [0.2324  0.2356  0.402   0.7     0.226  ]
% [0.1392  0.2039  0.10065 0.1693  0.3374 ]]

%%% Comments on the CLIP experiment in the Table
To validate the effectiveness of our disentanglement strategy, we performed an experiment where we edited attributes corresponding to smile, glasses, age, gender, and wearing a hat.
We edited samples using both the original and the disentangled directions while measuring the effect of each edit using CLIP~\cite{Radford2021CLIP} as a zero-shot classifier. 
We selected appropriate positive and negative prompts for each attribute. For smiling, glasses, and hat we used {\tt "A smiling person"}, {\tt "A person wearing glasses"} and {\tt "A person wearing a hat"} for the positive prompts respectively, and {\tt "A person"} as the negative prompt. 
For age and gender, we used {\tt "A man"} / {\tt "A woman"} and   {\tt "An old person" } /  {\tt "A young person"} respectively.
For each sample, we edited each of the five attributes and measured the change in attribute score according to CLIP.
Table~\ref{tab:disentanglement} shows the results. 
We can see that the original directions are highly entangled with other attributes while the disentangled directions induce the largest changes in the intended attributes. This demonstrates that semantic directions can be disentangled by a simple linear projection.

\section{Discussion and conclusion}
%% Punchline
We presented several supervised and unsupervised methods for finding interpretable directions in the recently proposed semantic latent space of Denoising Diffusion Models.
%% Unsupervised
We showed that the principal components in latent space correspond to global and semantically meaningful editing directions like pose, gender, and age. Additionally, we proposed a novel method for discovering directions based on a single input image. These directions correspond to highly localized changes in generated images, such as raising the eyebrows or opening/closing the mouth and eyes. 
% We further showed that 
Although these directions were found with respect to a specific image they can be transferred to different samples.

As our proposed methods enable high-quality editing of face images, we provide a broader impact statement in SM Sec.~\ref{sec:impact}.
Although our unsupervised approaches are effective in discovering meaningful semantics when the DDM was trained on aligned data like human faces, we found that models trained on less structured data have less interpretable principal directions. We refer the reader to SM Sec.~\ref{SM:pca-lsun} for experiments on models trained on churches and bedrooms. 
 
%% Supervised
Further, we proposed a conceptually simple supervised method utilizing the linear properties of the semantic latent space. 
We showed that a diverse set of face semantics can be revealed using an attribute classifier to annotate samples. 
Finally, we demonstrated that simple linear projection is an effective strategy for disentangling otherwise correlated semantic directions. 
%% Strengths in terms of not requiring retraining or adaptations  
All of our proposed methods apply to pretrained DDMs without requiring any adaptation to the model architecture, fine-tuning, optimization, or text-based guidance. 
Possible future avenues of our work include applications of the proposed approaches on different data domains. 

{\small
\bibliographystyle{ieee}
\bibliography{references}
}

\clearpage

\onecolumn
\appendix
% \subsection*{Supplemental Materials}

\begin{center}
{\Large \textbf{Supplemental Materials}}
\end{center}
\vspace{1cm}

\subsection{Illustration of \texorpdfstring{$h$}{h}-space.} \label{sm:hspace-diagram}

\begin{figure}[b] %[h!]
\centering
\includegraphics[width=0.6\linewidth]{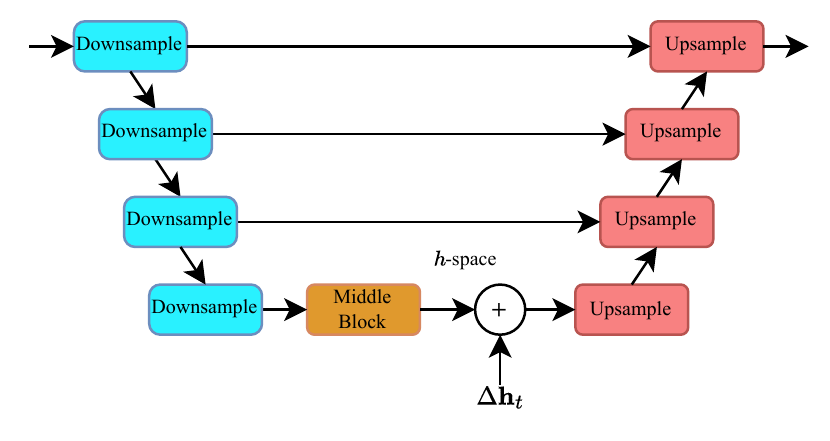}
\caption{\textbf{Illustration of $h$-space.}
In this paper, we define the semantic latent space of DDMs as the activation after the deepest bottleneck layer of the U-Net.
}
\label{fig:hspace}
\end{figure}

In this paper, we define $h$-space as the space of bottleneck activations $\mathbf{h}_t$ across each of the  $T$ timesteps in the synthesis process. See illustration in Fig.~\ref{fig:hspace}. Each downsampling block increases the number of channels while decreasing the spacial dimension of the feature maps. 
In our case, using the pretrained DDPM model trained on  CelebA released by Google\footnote{\url{https://huggingface.co/google/ddpm-ema-celebahq-256}}. The input pixel space has dimensions $(3,256,256)$ and the deepest feature map has dimensions $(512,8,8)$. Thus an element of $h$-space, $\mathbf{h}_{T:1}$, has dimensions $(T,512,8,8)$ and is defined as

% Mathematically, we define a latent code $\mathbf{h}_{T:1}$ in $h$-space as
\begin{align}
    \mathbf{h}_{T:1} = \mathbf{h}_{T} \otimes \mathbf{h}_{T-1}  \otimes \cdots  \otimes \mathbf{h}_{2} \otimes \mathbf{h}_{1}.
\end{align}

We apply directions in $h$ space by perturbing  $\mathbf{h}_{T:1}$ with some offset as $\mathbf{h}_{T:1} + \Delta\mathbf{h}_{T:1}$ during the generative process in \eqref{eq:ddim-reverse}. When $\eta_t \neq 0$ the clean image is completely specified by the triple $(\mathbf{x}_T, \mathbf{z}_{T:1}, \Delta\mathbf{h}_{T:1})$ and for $\eta_t = 0$ (DDIM) it is determined by the tuple $(\mathbf{x}_T, \Delta\mathbf{h}_{T:1})$. 

%\clearpage
\subsection{The effect of Asyrp}
\label{SM:asyrp}
\begin{figure}[ht]
\centering
\begin{subfigure}[b]{0.35\linewidth}
\includegraphics[width=\linewidth]{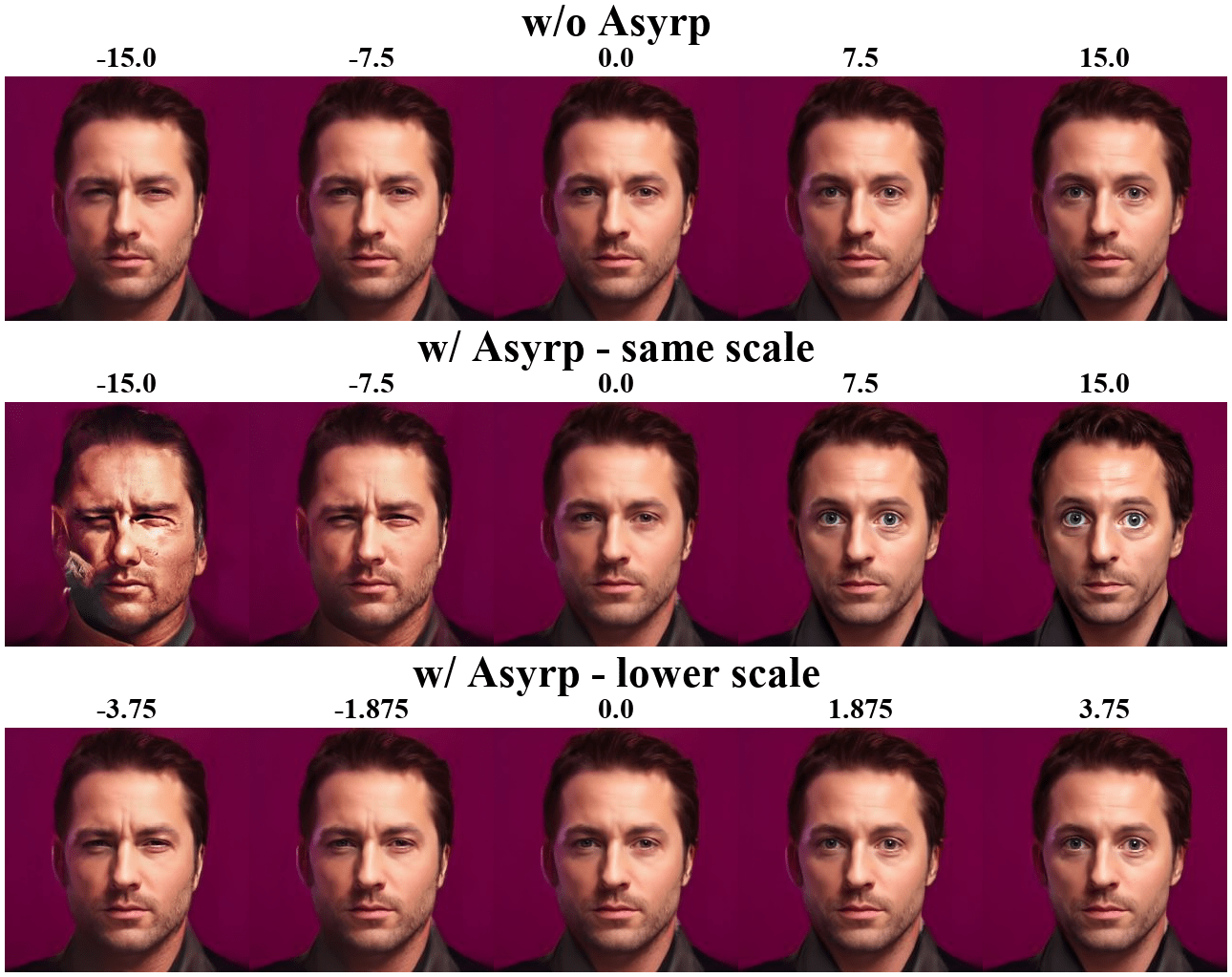}
\caption{Eyes}
\end{subfigure}
\begin{subfigure}[b]{0.35\linewidth}
\includegraphics[width=\linewidth]{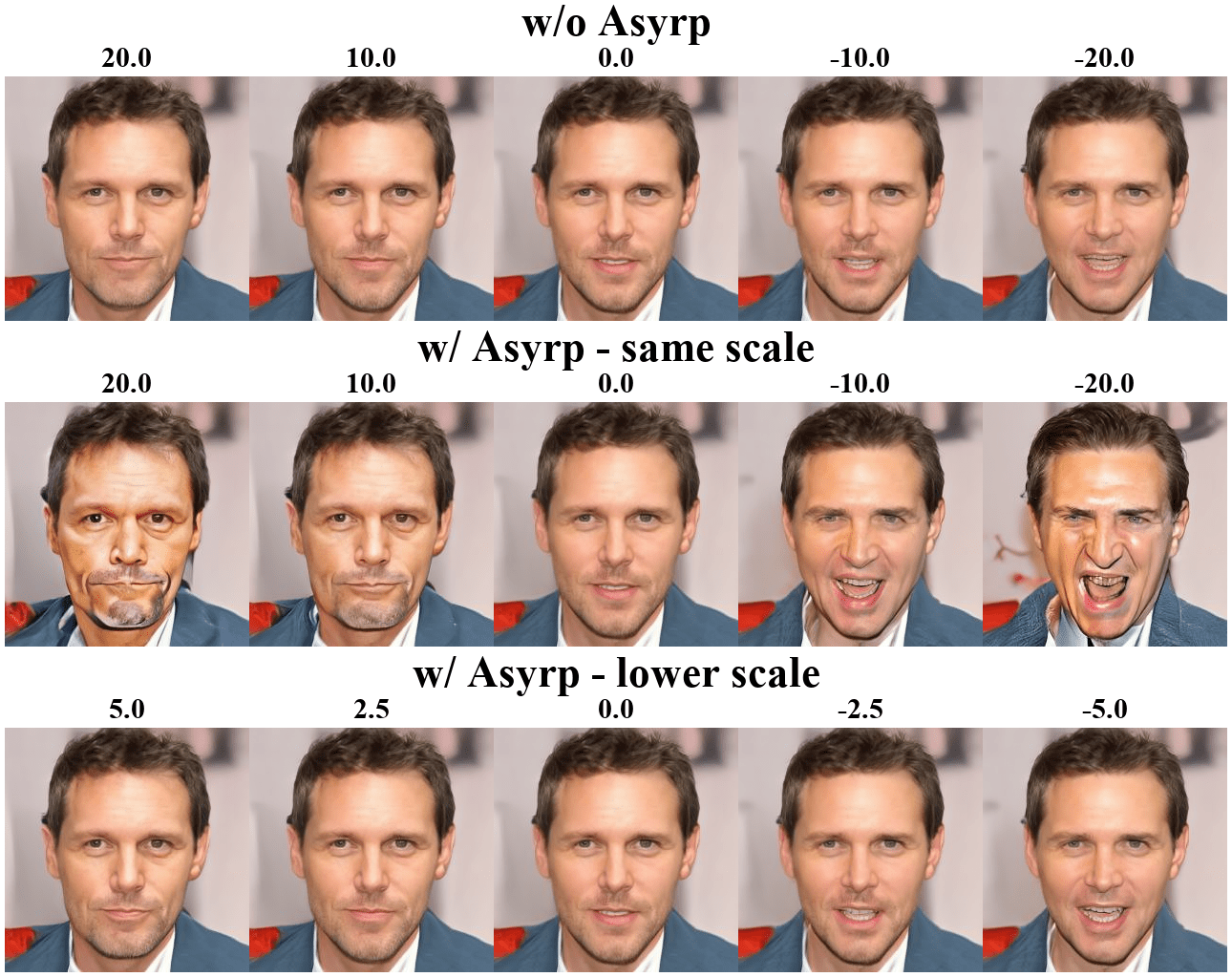}
\caption{Mouth}
\end{subfigure}
\caption{
\textbf{The Effect of Asyrp.} Results are shown for directions found with Alg.~\ref{alg:cap}.}
\label{SM:asyrp-plot1}
\end{figure}

\begin{figure}[ht]
\centering
% \begin{subfigure}[b]{0.45\linewidth}
% \includegraphics[width=\linewidth]{figs/sm/eyes1.png}
% \caption{Eyes}
% \end{subfigure}
% \begin{subfigure}[b]{0.45\linewidth}
% \includegraphics[width=\linewidth]{figs/sm/mouth1.png}
% \caption{Mouth}
% \end{subfigure}

\begin{subfigure}[b]{0.35\linewidth}
\includegraphics[width=\linewidth]{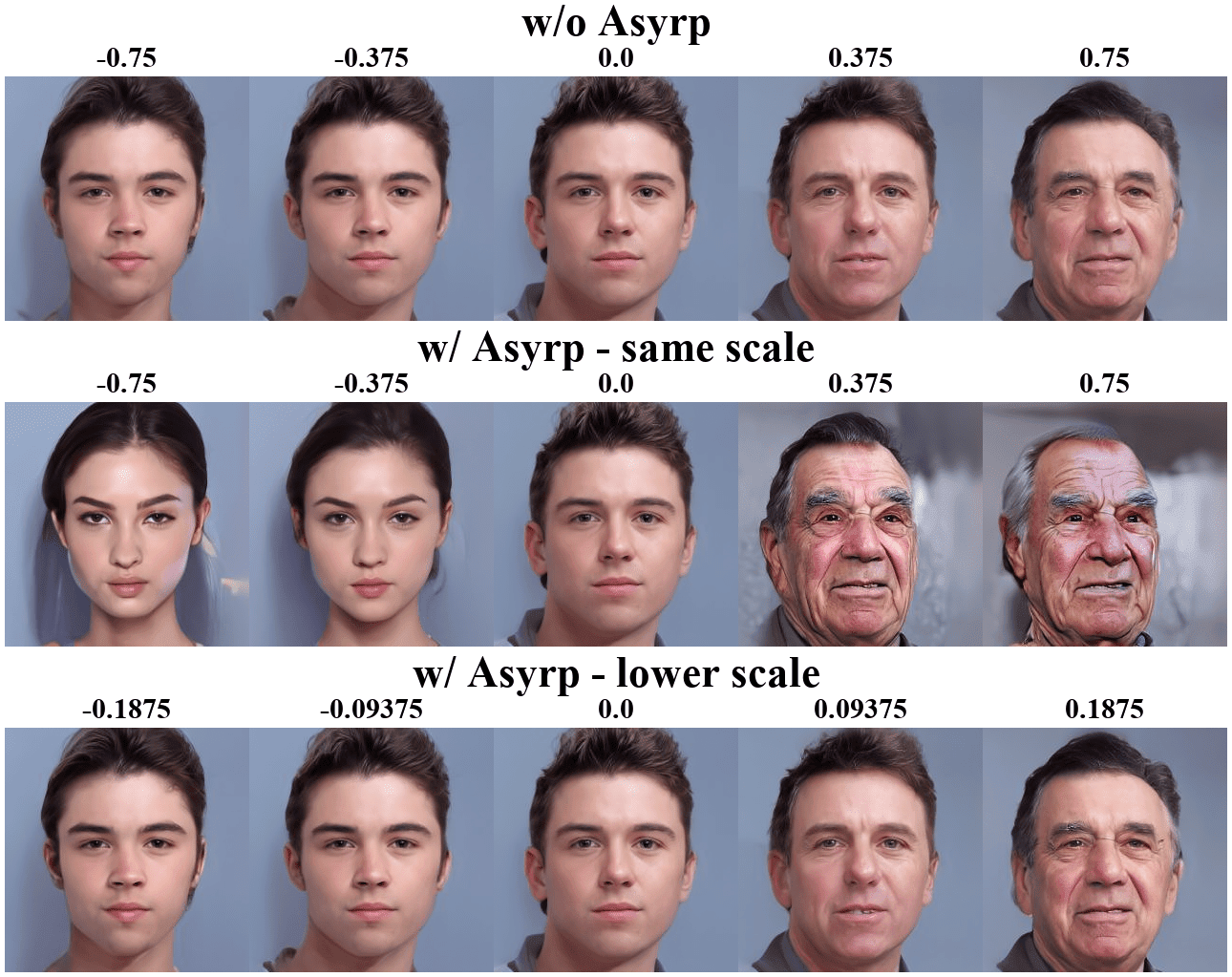}
\caption{Age}
\end{subfigure}
\begin{subfigure}[b]{0.35\linewidth}
\includegraphics[width=\linewidth]{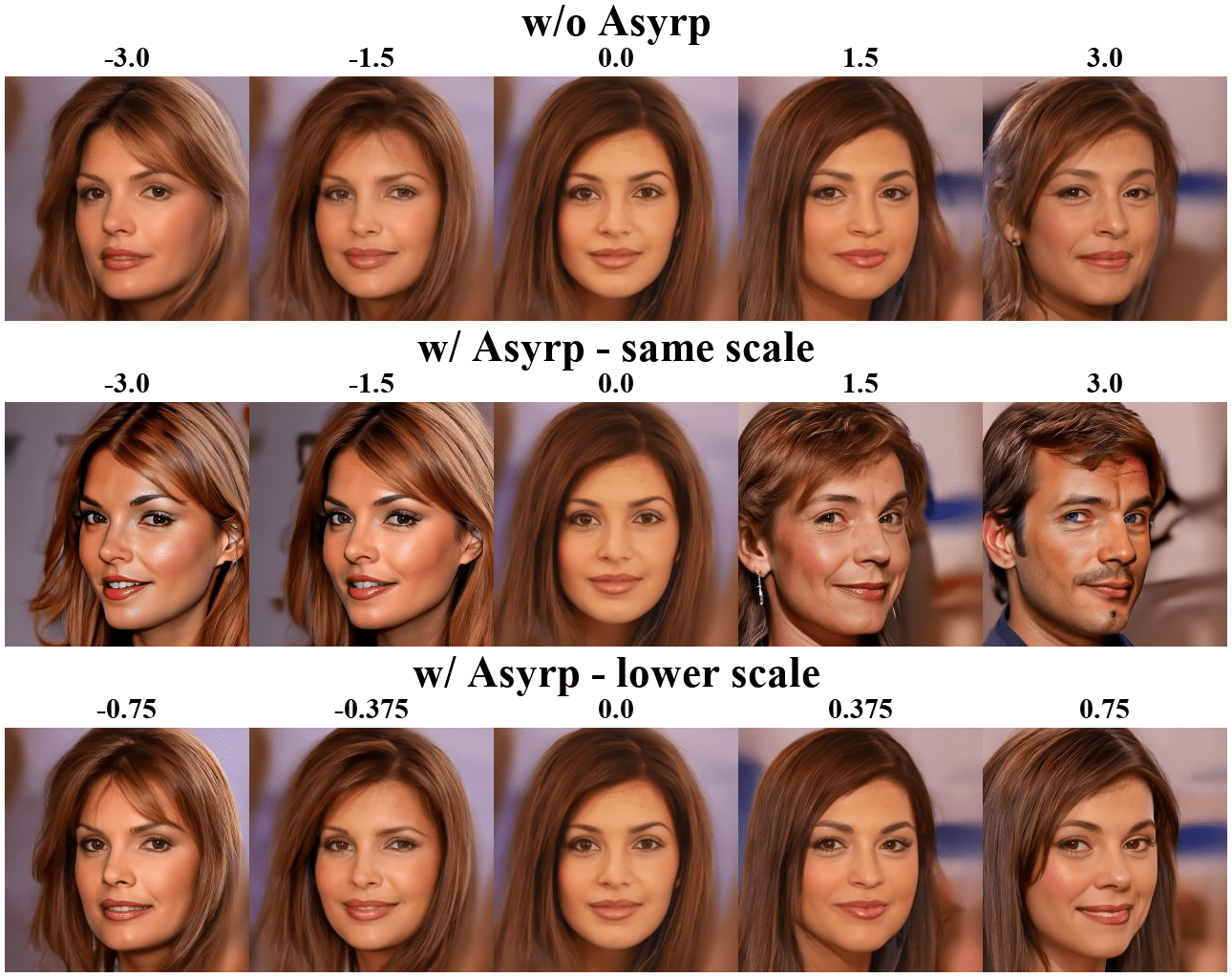}
\caption{Rotation}
\end{subfigure}

\begin{subfigure}[b]{0.35\linewidth}
\includegraphics[width=\linewidth]{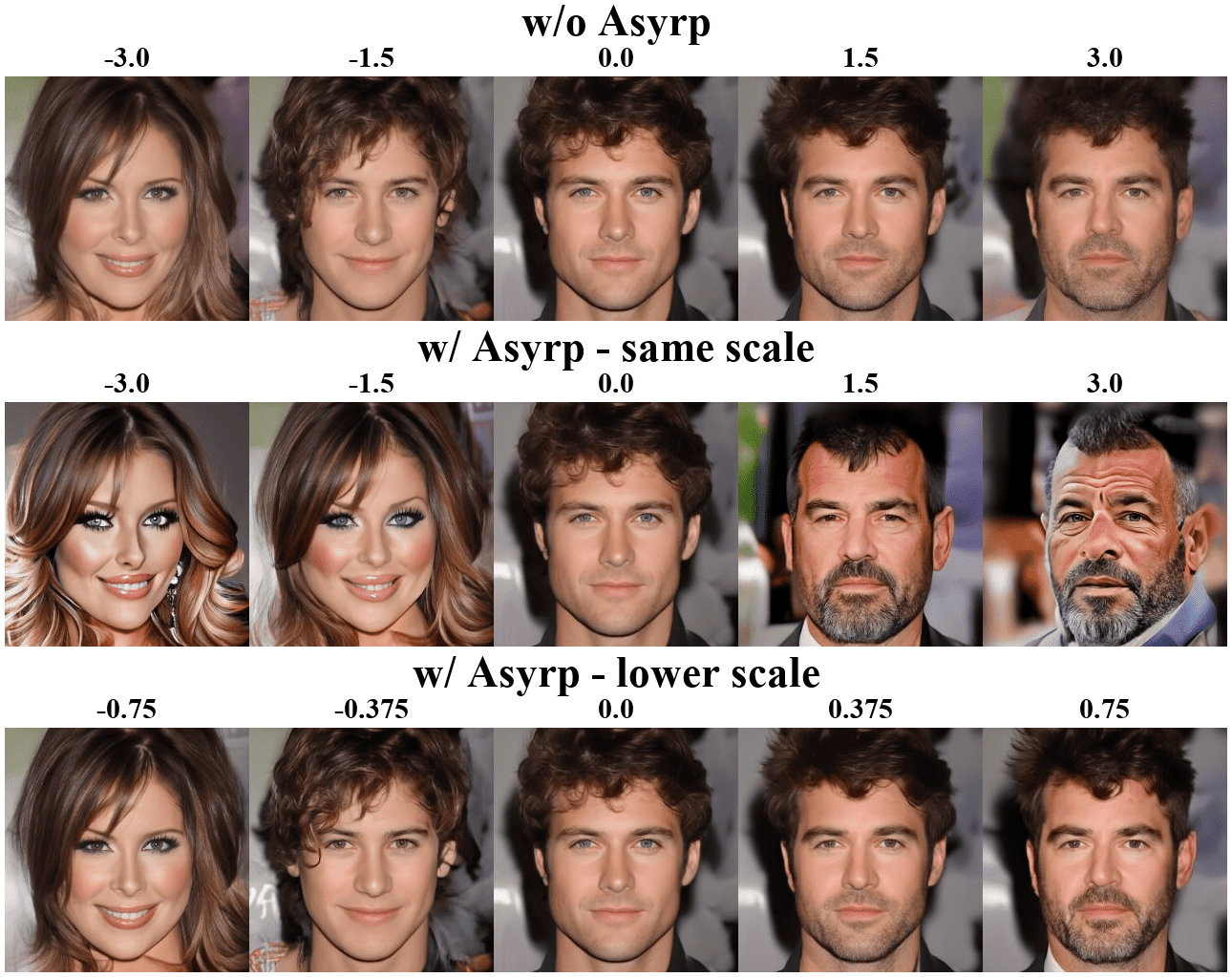}
\caption{Gender}
\end{subfigure}
\begin{subfigure}[b]{0.35\linewidth}
\includegraphics[width=\linewidth]{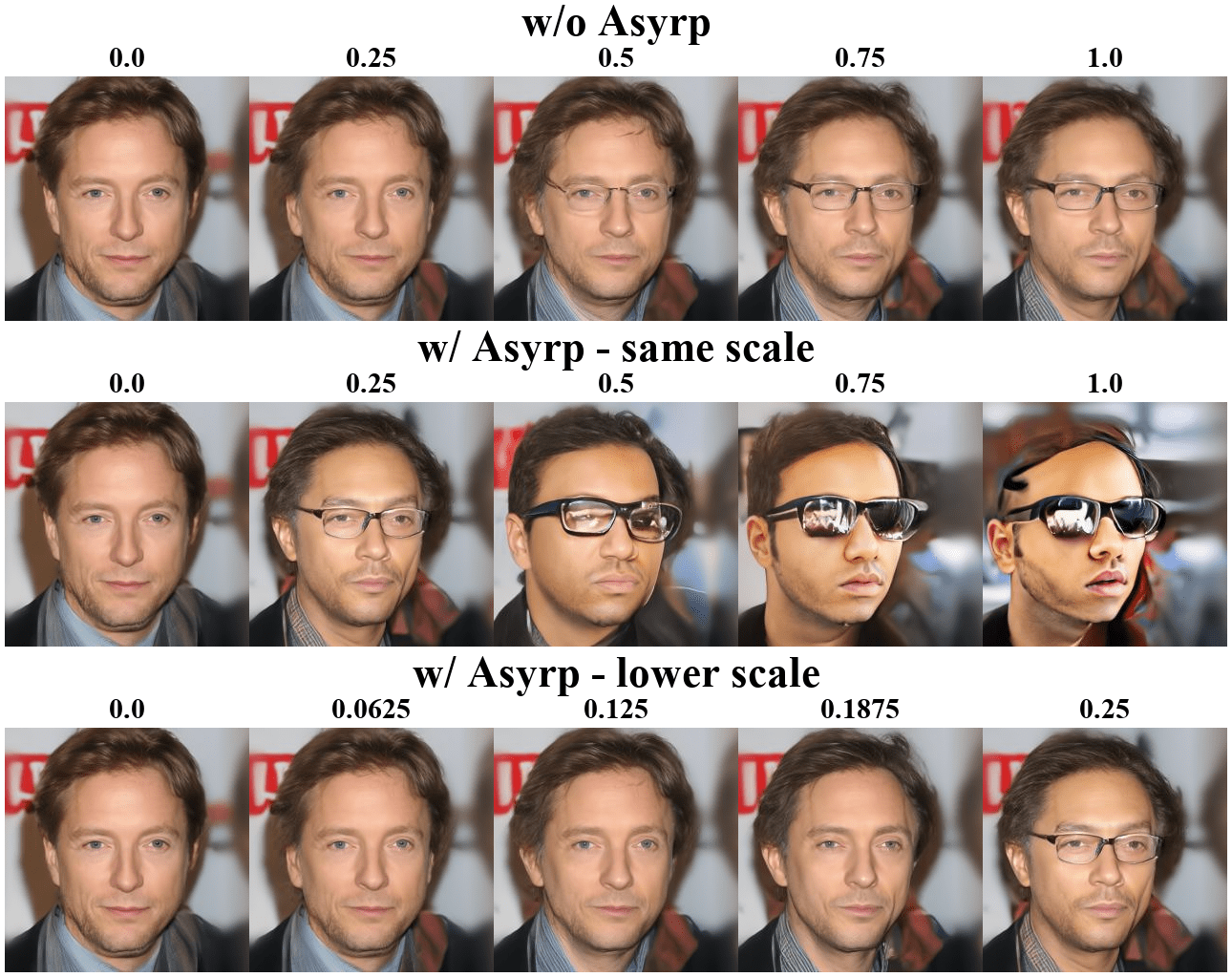}
\caption{Glasses}
\end{subfigure}

\caption{
\textbf{The effect of Asyrp.} Results are shown for directions found using the supervised method presented in Sec.~\ref{sec:Supervised methods}.
}
\label{SM:asyrp-plot2}
\end{figure}

In the main text, we stated that using Asyrp \cite{Kwon2022ddmhavesemantic} acts to amplify the effect edits in $h$-space. 
However, Asyrp is computationally costly since it requires two forward passes of the U-Net at each denoising step. 
Hence, Asyrp is not used for any of the results shown in the main paper.
% as similar edits can be achieved by simply increasing the scale.
In Figs.~\ref{SM:asyrp-plot1} and \ref{SM:asyrp-plot2} we qualitatively compare edits with and without using Asyrp. We observe that simply adjusting the scale of the applied direction results in very similar edits.

%\clearpage

% and classifier annotation respectively.

%%%%%%%%%%%%%%%%%%%%%%%%%%%%%%%%%%%%%%%%%%%%%%

\clearpage
\subsection{A Note on image-specific directions}\label{SM:noteonnoiseprediction}
In the main paper, we state that the right singular vectors of the Jacobian of $\bm{\epsilon}_t^\theta$ with respect to $h$-space, denoted as $\mathbf{J}_t$, are the set of orthogonal vectors in $h$-space which perturb the noise prediction $\bm{\epsilon}_t^\theta$ the most.
An equivalent statement is that those right singular vectors perturb the predicted image $\mathbf{P}_t(\mathbf{x}_t ,\mathbf{h}_t)$ at timestep $t$ the most. 
Specifically, since
\begin{equation}
\mathbf{P}_t(\mathbf{x}_t ,\mathbf{h}_t) =
\frac{\mathbf{x}_t - \sqrt{1-\alpha_t}}{\sqrt{\alpha_t} }
\bm{\epsilon}^\theta_t(\mathbf{x}_t,\mathbf{h}_t)
\end{equation}
 we have that 
\begin{align}
\frac{\partial}{\partial \mathbf{h}_t} \mathbf{P}_t(\mathbf{x}_t ,\mathbf{h}_t) 
&= -\frac{\sqrt{1-\alpha_t}}{\sqrt{\alpha_t}} \frac{\partial}{\partial \mathbf{h}_t} \\
\bm{\epsilon}^\theta_t(\mathbf{x}_t,\mathbf{h}_t)  &= -\frac{\sqrt{1-\alpha_t}}{\sqrt{\alpha_t}} \mathbf{J}_t.
% \frac{\mathbf{x}_t - \sqrt{1-\alpha_t}  }{\sqrt{\alpha_t}}
\end{align}
Thus, the eigenvectors  of $(\partial \mathbf{P}_t/\partial \mathbf{h}_t)^\mathrm{T}(\partial \mathbf{P}_t/\partial \mathbf{h}_t)$ and  $\mathbf{J}_t^\mathrm{T}\mathbf{J}_t$ are the same with the same ordering.

\subsection{Image-specific directions at different timesteps} \label{SM:jacobiantimesteps}

\begin{figure*}[ht]
\centering
\includegraphics[width=0.95\linewidth]{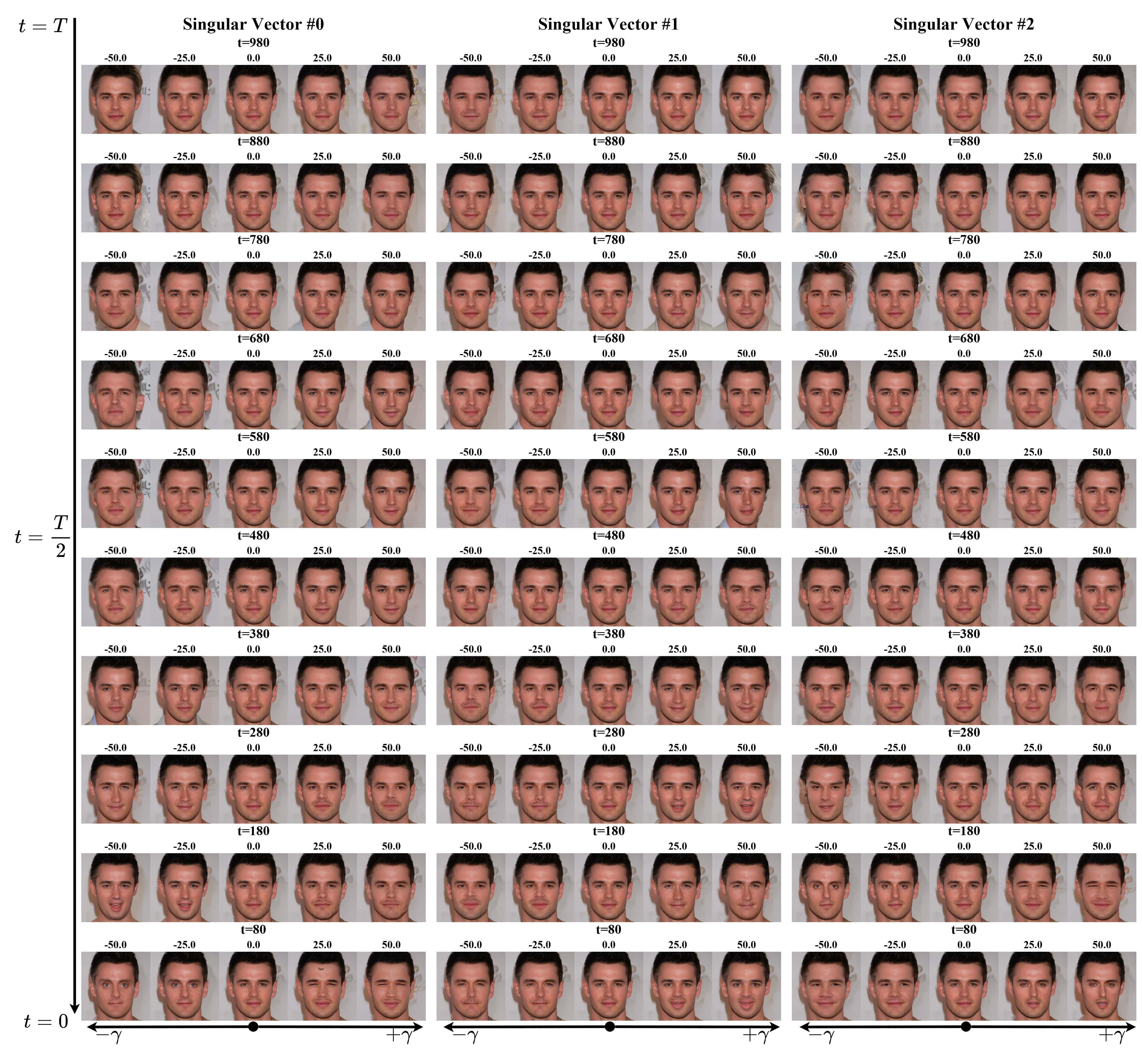}
\caption{\textbf{Directions found by Alg.~\ref{alg:cap}.}}
\label{SM:poweriter-seed199805}
\end{figure*}

\begin{figure*}[ht]
\centering
\includegraphics[width=0.95\linewidth]{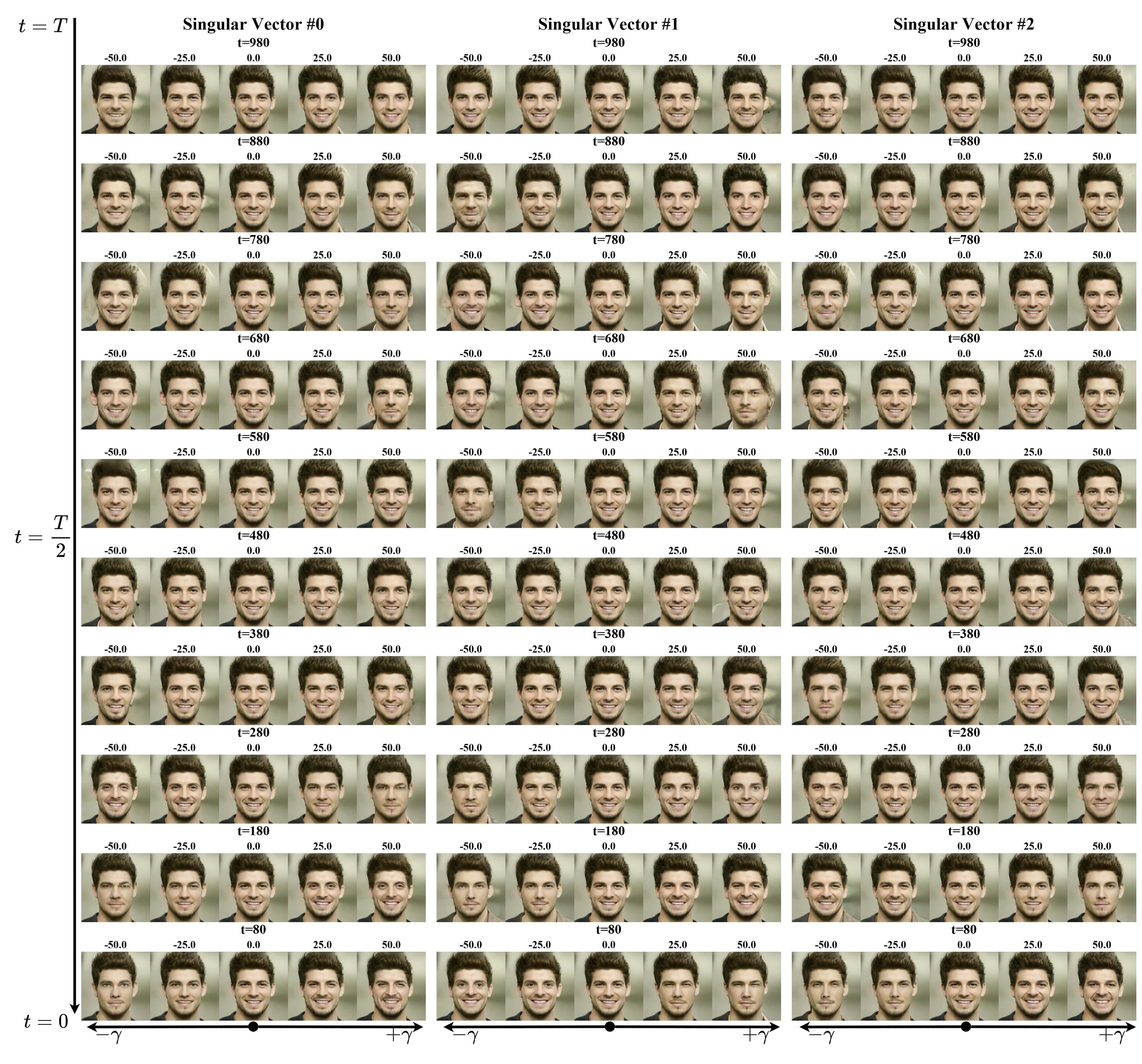}
\caption{\textbf{Directions found by Alg.~\ref{alg:cap}.}}
\label{SM:poweriter-seed445314}
\end{figure*}

\begin{figure*}[ht]
\centering
\includegraphics[width=0.95\linewidth]{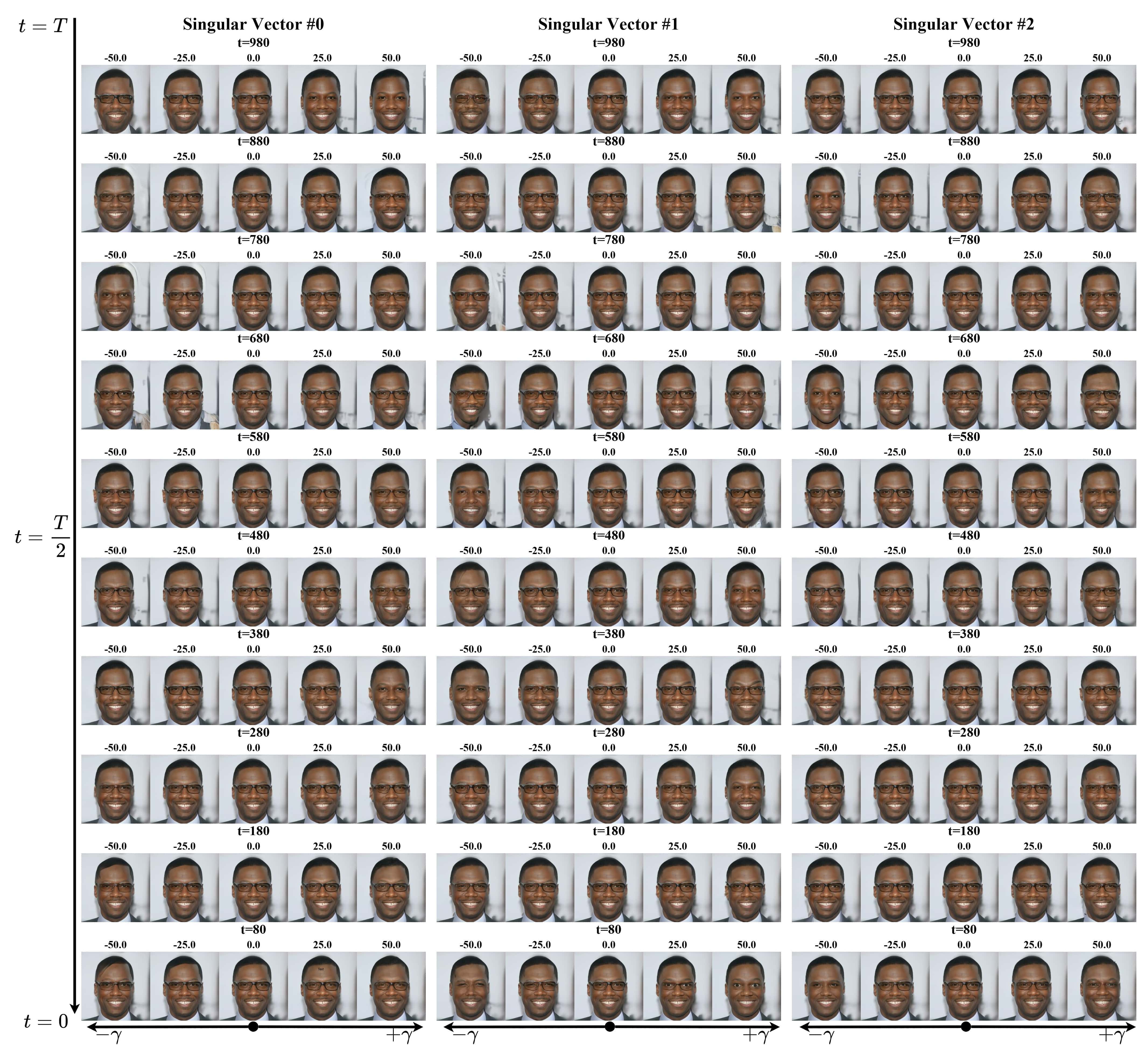}
\caption{\textbf{Directions found by Alg.~\ref{alg:cap}.}}
\label{SM:poweriter-seed655092}
\end{figure*}

\begin{figure*}[ht]
\centering
\includegraphics[width=0.95\linewidth]{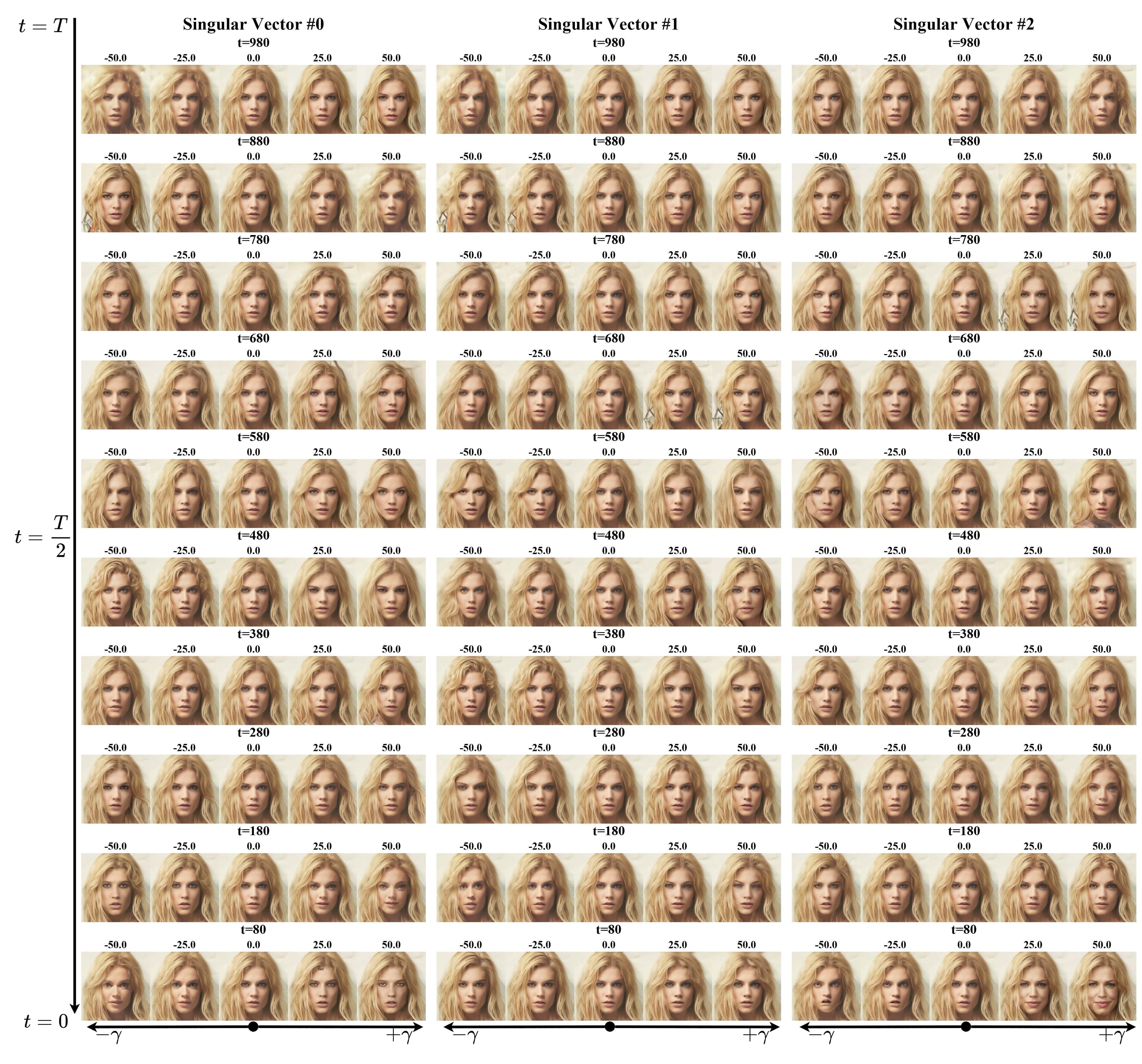}
\caption{\textbf{Directions found by Alg.~\ref{alg:cap}.}}
\label{SM:poweriter-seed825356}
\end{figure*}

Our proposed image-specific unsupervised method in Alg.~\ref{alg:cap} finds different directions for each timestep. 
In Figures \ref{SM:poweriter-seed199805}, \ref{SM:poweriter-seed445314}, \ref{SM:poweriter-seed655092} and \ref{SM:poweriter-seed825356} we show the effect of the three dominant directions (the three top singular vectors of the Jacobian) at different timesteps along the reverse diffusion process. 

%%%%%%%%%%%%%%%%%%%%%%%%%%%%%%%%%%%%%%%%%%%%%%

\clearpage
\subsection{Sequential algorithm for Jacobian subspace iteration}\label{SM:jacobian}%~%

As mentioned in the main text, Alg.~\ref{alg:cap} can be memory intensive when calculating a large number of singular vectors in parallel. 
In cases where limited memory is available, we provide an alternative sequential version of our method in Alg.~\ref{alg:jacsubspace_sequential}. 
Here we calculate the singular values and vectors in mini-batches of size $ b $.  The value of $ b $ should be set according to the parallel computation capacity. For example, in the special case of $ b = 1 $, the algorithm computes the vectors one by one and will use small memory. Note that lowering the mini-batch size $b$ comes at the expense of longer running time.

% This is efficiently done by utilizing forward mode automatic differentiation.
% Further \eqref{eq:JacobianVectorProd} can be calculated in parallel for multiple vectors using the batched Jacobian-vector product \eg in Pytorch. 

\begin{algorithm}[ht]
\caption{Sequential Jacobian subspace iteration}\label{alg:jacsubspace_sequential}
\begin{algorithmic}
\Require function to differentiate $ \mathbf{f} : \mathbb{R}^{d_{\text{in}}} \to \mathbb{R}^{d_{\text{out}}}$, point at which to differentiate
$\mathbf{h} \in  \mathbb{R}^{d_{\text{in}}}$, initial guess $\mathbf{\Theta} \in  \mathbb{R}^{d_{\text{in}} \times k} $ [optional],
mini-batch size $ b<k $ 
\Ensure $ (\mathbf{U}, \mathbf{\Sigma}, \mathbf{V}^\mathrm{T}) $ -- $k$ top singular values and vectors of the Jacobian $ {\partial \mathbf{f} }/{ \partial \mathbf{h}}$
\State  \textbf{Initialization: } $
\mathbf{y} \gets \mathbf{f}(\mathbf{h}), \
i_{\text{start}} \gets 1, \
i_{\text{end}} \gets b, \
\mathbf{V} \gets [ \ ] , \
\mathbf{\Sigma} \gets [ \ ] , \
\mathbf{U} \gets [ \ ]  $
\While{$i_{\text{start}} \leq k$}
\If{$\mathbf{\Theta}$ is empty}
    \State $\mathbf{\Phi} \gets $ i.i.d.\@ standard Gaussian samples in $ \mathbb{R}^{d_{\text{in}}\times (i_{\text{end}}-i_{\text{start}}+1) } $
\Else
    \State $\mathbf{\Phi} \gets $ columns $i_{\text{start}}$ to $i_{\text{end}}$ of $\mathbf{\Theta}$
\EndIf
\State $  \mathbf{Q},\mathbf{R} \gets \mathrm{QR}(\mathbf{\Phi}) $
\Comment{Reduced QR decomposition}
\State $\mathbf{\Phi} \gets \mathbf{Q}$
\Comment{Ensures $ \mathbf{\Phi}^\mathrm{T} \mathbf{\Phi} = \mathbf{I} $}
\While{stopping criterion}
\If{$\mathbf{V}$ is not empty}
\State $\mathbf{\Phi} \gets \left[\mathbf{I} -  \mathbf{V}\left(\mathbf{V}^\mathrm{T}\mathbf{V}\right)^{-1} \mathbf{V}^\mathrm{T} \right] \mathbf{\Phi} $
\State $  \mathbf{\Phi},\mathbf{R} \gets \mathrm{QR}(\mathbf{\Phi}) $
\Comment{Reduced QR decomposition}
\EndIf
\State $\mathbf{\Psi} \gets \partial \mathbf{f} ( \mathbf{h}+a \mathbf{\Phi} ) / \partial a $ at $ a = 0$
\Comment{Batch forward}
\State $\hat{\mathbf{\Phi}} \gets \partial (\mathbf{\Psi}^\mathrm{T}\mathbf{ y })/\partial \mathbf{h}$
\State $\mathbf{\Phi},\mathbf{S}, \mathbf{R} \gets \mathrm{SVD}(\hat{\mathbf{\Phi}})$
\Comment{Reduced SVD}
\EndWhile
\State $\mathbf{V} \gets [\mathbf{V} ; \mathbf{\Phi}]$
\State\vspace*{-\baselineskip}
    \begin{fleqn}[\dimexpr(\leftmargini-\labelsep)]
        \setlength\belowdisplayskip{3pt}
        \setlength\abovedisplayskip{3pt}
        \begin{equation*}
            \mathbf{\Sigma} \gets
            \begin{bmatrix}
                \mathbf{\Sigma} & \mathbf{0} \\
                \mathbf{0}   &   \mathbf{S}^{1/2} 
            \end{bmatrix}
        \end{equation*}
    \end{fleqn}%
\State $\mathbf{U} \gets [\mathbf{U} ; \mathbf{\Psi}]$
\State  $i_{\text{start}} \gets i_{\text{start}}+b $
\State  $i_{\text{end}} \gets \min\{ i_{\text{end}}+b,k\} $

\EndWhile
\State Orthonormalize $\mathbf{U}$
%\State $\mathbf{\Sigma} \gets (\mathbf{\Sigma}^2)^{1/2} $

% \State $\mathbf{y} \gets \mathbf{f}(\mathbf{h})$ % , \;  a \gets 1  $
% \If{$\mathbf{V}$ is empty}
%     \State $\mathbf{V} \gets $ i.i.d.\@ standard Gaussian samples
% \EndIf
% \State $  \mathbf{Q},\mathbf{R} \gets \mathrm{QR}(\mathbf{V}) $
% \Comment{Reduced QR decomposition}
% \State $\mathbf{V} \gets \mathbf{Q}$
% \Comment{Ensures $ \mathbf{V}^\mathrm{T} \mathbf{V} = \mathbf{I} $}
% \While{stopping criteria}
% % \State $\mathbf{B} \gets \mathbf{h}-\mathbf{U} $ 
% % \Comment{$\mathbf{h}$ broadcasted}
% \State $\mathbf{U} \gets \partial \mathbf{f} ( \mathbf{h}+a \mathbf{V} ) / \partial a $ at $ a = 0$
% \Comment{Batch forward}
% \State $\hat{\mathbf{V}} \gets \partial (\mathbf{U}^\mathrm{T}\mathbf{ y })/\partial \mathbf{h}$
% \State $\mathbf{V},\mathbf{\Sigma^2}, \mathbf{R} \gets \mathrm{SVD}(\hat{\mathbf{V}})$
% \Comment{Reduced SVD}
\end{algorithmic}
\end{algorithm}

\clearpage
\subsection{Facial expressions from real data.}\label{SM:bu3dfe}
\begin{figure}[h]
\centering
\includegraphics[width=0.7\linewidth]{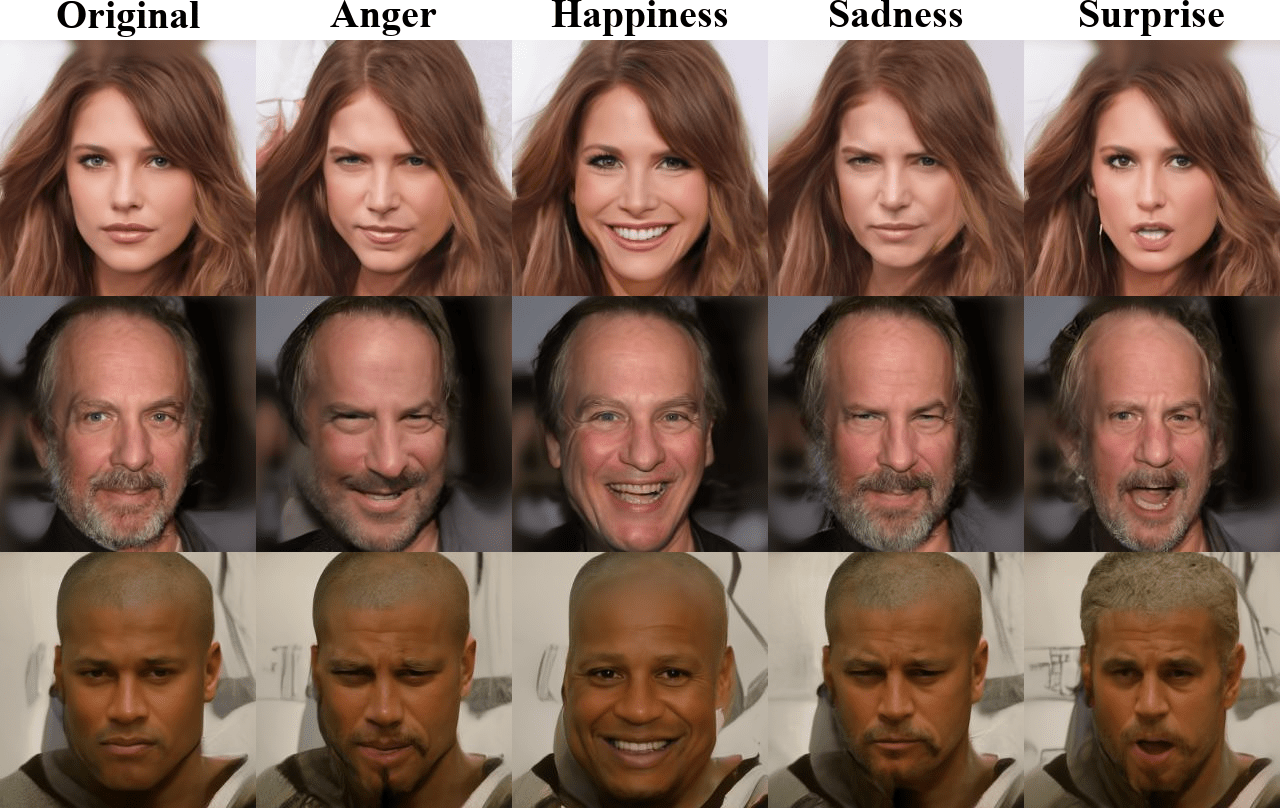}
\caption{
\textbf{Facial expressions from real data.}
We extract semantic directions corresponding 
to different facial expressions using a data set of real images. 
The directions are calculated via DDIM inversion and applied in the semantic $h$-space to synthetic images.
}
\label{fig:bu3dfe}
\end{figure}

We conducted an additional experiment where domain-specific semantic directions were extracted using real images as supervision. We wish to find directions corresponding to expressions like happiness, sadness, and surprise. Here we used the BU3DFE data set~\cite{yin2006bu3dfe}.
BU3DFE contains real images of $100$ subjects, each performing a neutral expression in addition to each of the prototypical facial expressions at various intensity levels.
Using DDIM inversion ($\eta_t = 0$) we recorded $\mathbf{h}_{T:1}$ during the inversion process and used \eqref{eq:linear_direction} to calculate directions. 
We used the most intense expressions for the positive examples and the neutral expressions for the negative examples. 
The effect of the directions found using our method is shown in Fig.~\ref{fig:bu3dfe}. The extracted directions are shown on generated samples.  The figure shows that latent directions in $h$-space can successfully be found by applying our supervised method presented in Sec.~\ref{sec:supervised} on a dataset of real images.
 %bu3dfe

\subsection{Broader impact}\label{sec:impact}

In this paper, we have introduced several techniques for semantic editing of human faces using DDMs. While the creation of high-quality edited images that are difficult to distinguish from real images has significant positive applications, there is also the potential for malicious or misleading use, such as in the creation of deepfakes. 
Although some research has focused on detecting and mitigating the risk of AI-edited images, these have mostly focused on GANs \cite{Wang2022GANgeneratedFD} and, so far, there has been little research into detecting images that have been edited using DDMs. Given the differences in the generative process between DDMs and GANs, methods which are effective in detecting images edited by GANs might not be as effective for images edited by DDMs \cite{meng2022sdedit}.
Further research is needed to develop effective methods for forensic analysis of edits using DDMs. 
Such research could help address the risk of malicious use of image-editing technologies.

%%%%%%%%%%%%%%%%%%%%%%%%%%%%%%%%%%%%%%%%%%%%%%
\clearpage
\subsection{Unsupervised methods on other domains}
\label{SM:pca-lsun}

\begin{figure*}[ht]
\centering
\includegraphics[width=0.75\textwidth]{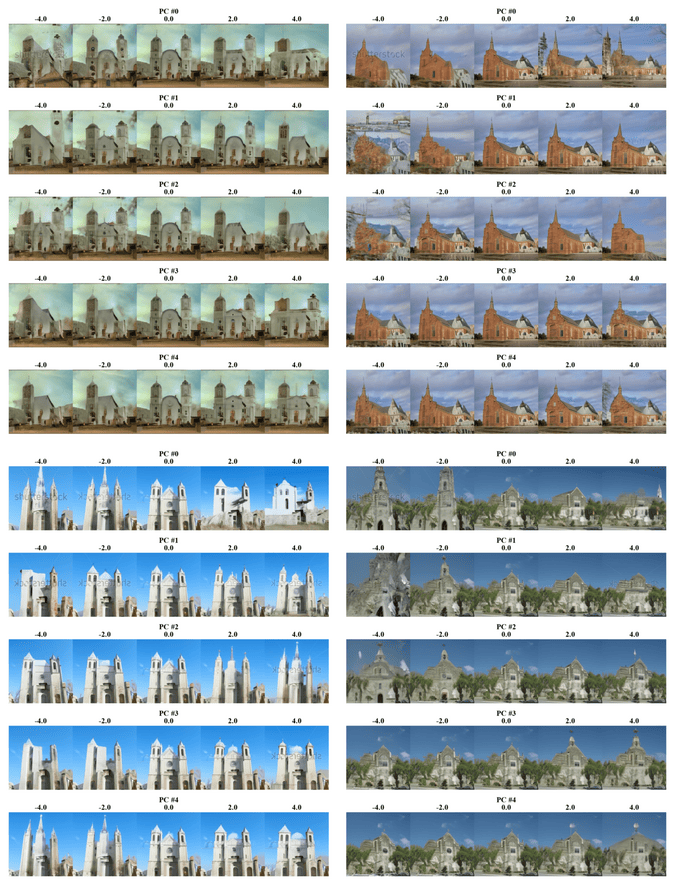}
\caption{
\textbf{PCA directions.}
For a DDM trained on churches.}
\label{SM:pca-church}
\end{figure*}

\begin{figure*}[ht]
\centering
\includegraphics[width=0.75\textwidth]{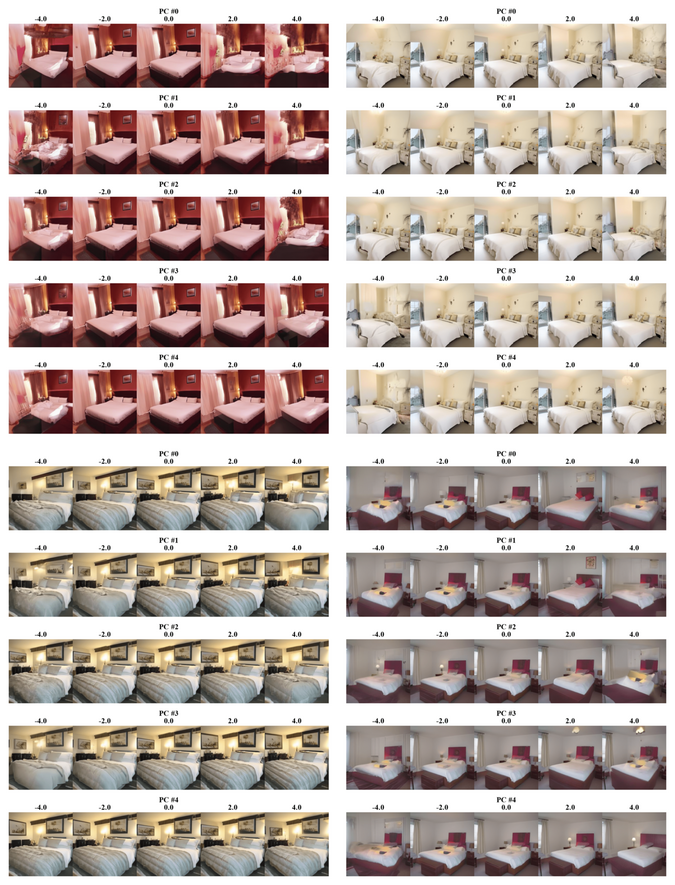}
\caption{
\textbf{PCA directions.}
For a DDM trained on bedrooms.}
\label{SM:pca-bedrooms}
\end{figure*}

\begin{figure*}[ht]
\centering
\includegraphics[width=0.9\textwidth]{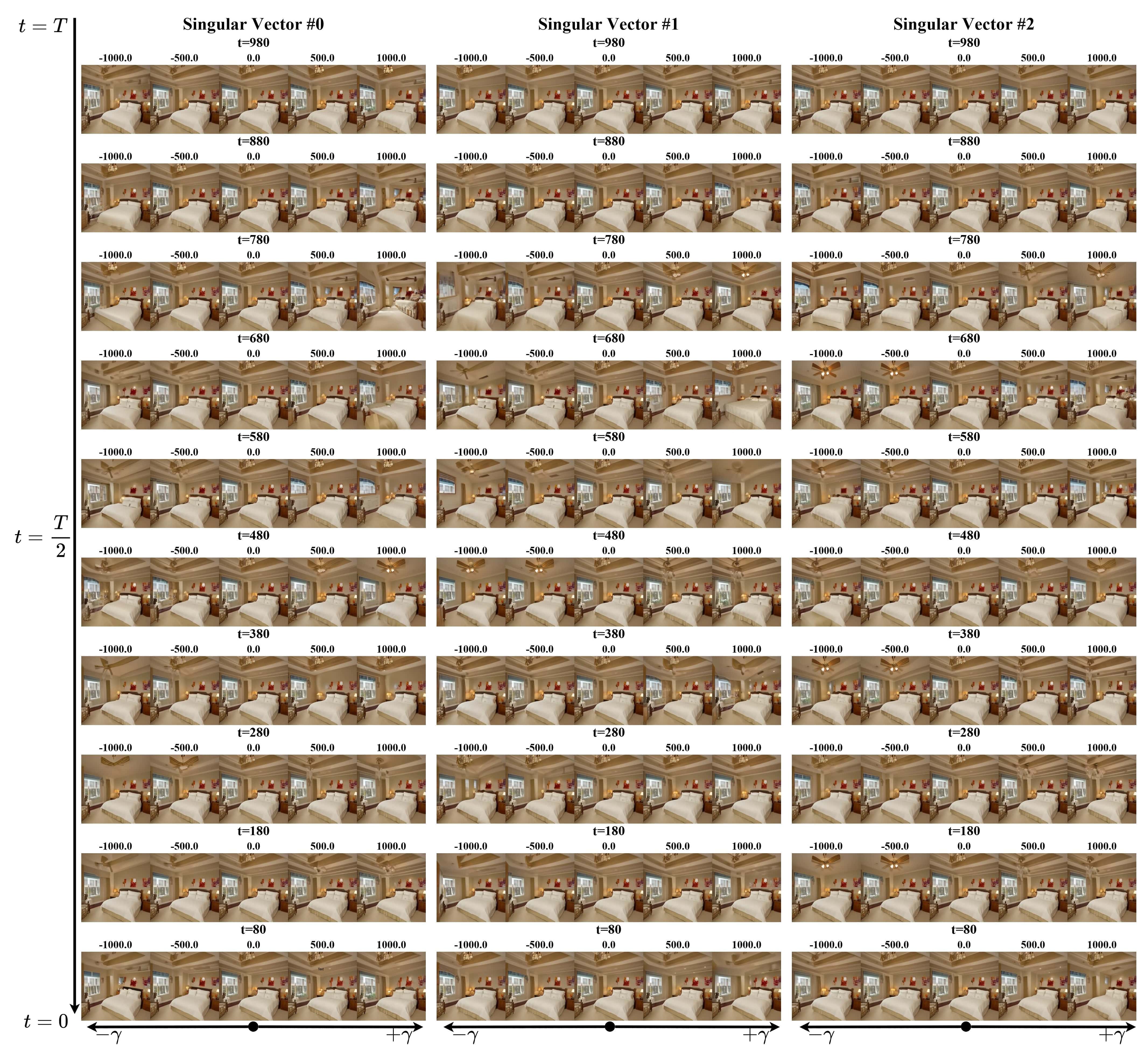}
\caption{
\textbf{Directions found with Alg.~\ref{alg:cap}}.
For a DDM trained on bedrooms.}
\label{SM:poweriter-bedrooms}
\end{figure*}

\begin{figure*}[ht]
\centering
\includegraphics[width=0.9\textwidth]{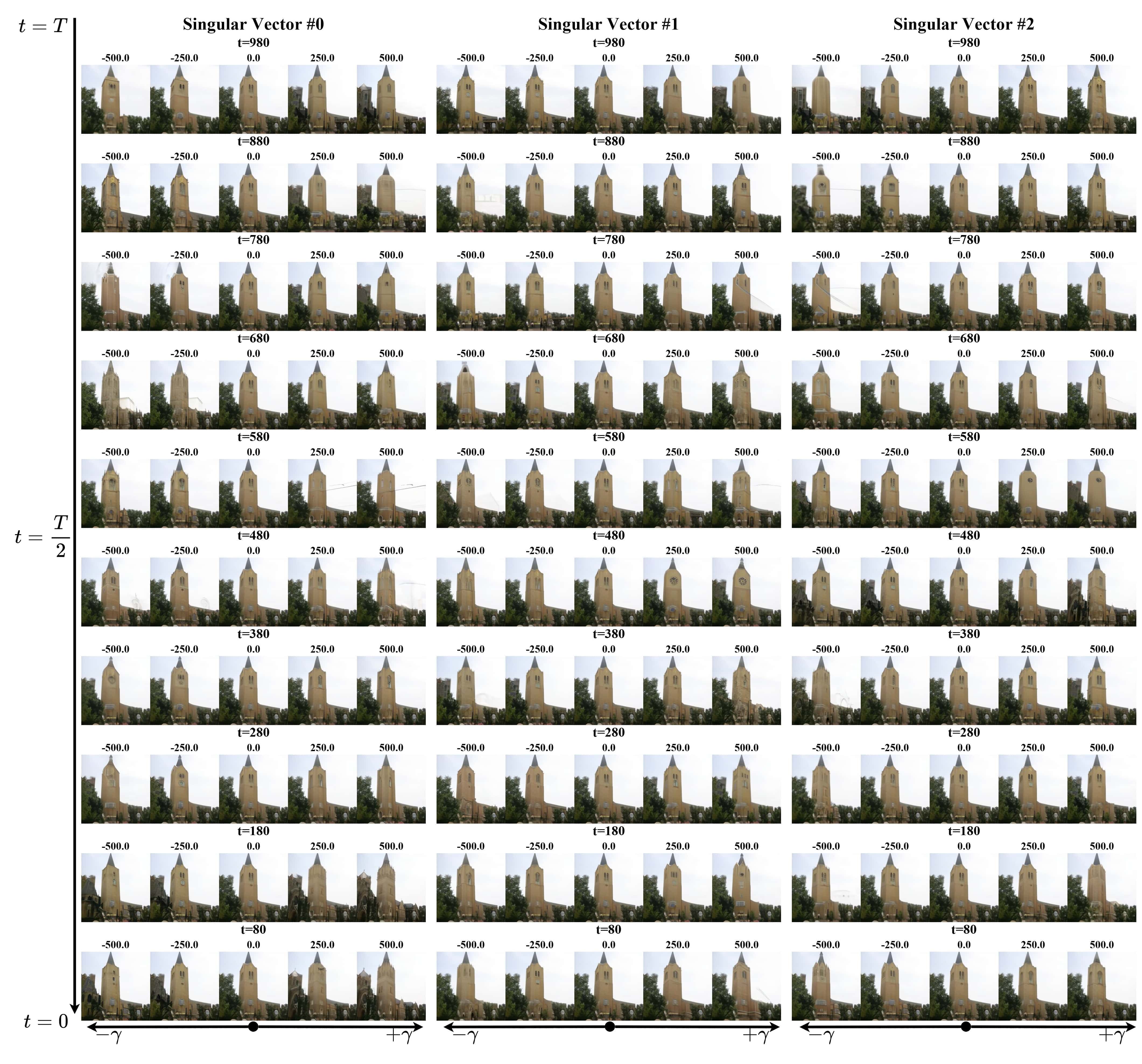}
\caption{
\textbf{Directions found with Alg.~\ref{alg:cap}}.
For a DDM trained on churches.}
\label{SM:poweriter-churches}
\end{figure*}

In addition to the model\footnote{\url{https://huggingface.co/google/ddpm-ema-celebahq-256}} trained on CelebA, which is used throughout the main paper, we also conducted experiments with models trained on 
churches\footnote{\url{https://huggingface.co/google/ddpm-ema-church-256}} 
and bedrooms\footnote{\url{https://huggingface.co/google/ddpm-ema-bedroom-256}}.
Although the unsupervised directions found with both PCA and Alg.~\ref{alg:cap}
on these models lead to various changes to the images, these directions are less interpretable than those obtained for faces in the main paper. 
We showcase the first $5$ PCA directions on the models trained on churches and bedrooms in Figures~\ref{SM:pca-church} and \ref{SM:pca-bedrooms} and directions found using 
Alg.~\ref{alg:cap} in Figures~\ref{SM:poweriter-churches} and \ref{SM:poweriter-bedrooms}.

% \clearpage
% \subsection{Transferability of semantic directions}

% \begin{figure}[t]
% \centering
% \includegraphics[width=\linewidth]{figs/poweriter/poweriterfig3_annotated.png}
% \caption{
% \textbf{Unsupervised image-specific edits.}
% Spectral analysis of the Jacobian of $\bm{\epsilon}_t^\theta$ yields directions corresponding to localized changes in the generated image, \eg eyes opening/closing and raising of the eyebrows.
% Although this method is image-specific, directions found for one sample can be transferred to others, where they result in semantically similar edits. }
% \label{fig:poweriter}
% \end{figure}

% \begin{figure}[tb]
% \centering
% \includegraphics[width=\textwidth]{figs/pca/pixel-pca_test_all50steps250samples-eta1.jpg}
% \caption{Pixel CelebA PCA 50 inference steps 250 samples}
% \end{figure}

\end{document}